\documentclass[11pt]{article}
\usepackage{float}
\usepackage[final]{acl}

\usepackage{times}
\usepackage{latexsym}

\usepackage[T1]{fontenc}

\usepackage[utf8]{inputenc}

\usepackage{microtype}

\usepackage{inconsolata}

\usepackage{graphicx}
\usepackage{subcaption}

\usepackage{booktabs}
\usepackage{multirow}

\usepackage{graphicx}
\usepackage{tikz}

\usepackage[most]{tcolorbox}
\tcbuselibrary{listings,skins,breakable}
\usepackage{xcolor}
\usepackage{listings}
\usepackage{textcomp}

\usepackage[table]{xcolor}

\lstdefinestyle{prompttext}{
    basicstyle=\ttfamily\scriptsize,
    breaklines=true,
    breakatwhitespace=false,
    columns=fullflexible,
    keepspaces=true,
    showstringspaces=false,
    upquote=true,
    escapeinside={(*@}{@*)},
    literate=
      {→}{{$\rightarrow$}}1
      {—}{{---}}1
      {–}{{--}}1
      {“}{{``}}1
      {”}{{''}}1
      {’}{{'}}1
}

\newtcblisting{widepromptbox}[2][]{
    enhanced jigsaw,
    breakable,
    listing only,
    listing engine=listings,
    listing options={style=prompttext},
    colback=gray!3,
    colframe=black!70,
    coltitle=white,
    colbacktitle=black!75,
    title=\textbf{#2},
    fonttitle=\small,
    arc=1.5mm,
    boxrule=0.6pt,
    left=2mm,
    right=2mm,
    top=1mm,
    bottom=1mm,
    width=0.96\textwidth,
    #1
}

\title{\textsc{MineExplorer}: Evaluating Open-World Exploration of\\ MLLM Agents in Minecraft}

\author{Tianjie Ju\textsuperscript{1,2}\thanks{Work completed while Tianjie Ju and Zheng Wu were interns at Meituan. $^\dag$ Corresponding authors. This work was supported by the Shanghai Municipal Science and Technology Major Project (2025SHZDZX025G08).}\quad
Yueqing Sun\textsuperscript{2}\quad
Zheng Wu\textsuperscript{1,2}\quad
Wei Zhang\textsuperscript{2}\quad
Yaqi Huo\textsuperscript{2}\quad
Xi Su\textsuperscript{2}\\
{\bf Qi Gu\textsuperscript{2}$^\dag$\quad
Xunliang Cai\textsuperscript{2}\quad
Gongshen Liu\textsuperscript{1}\quad
Zhuosheng Zhang\textsuperscript{1$^\dag$}
}\\
\textsuperscript{1}School of Computer Science, Shanghai Jiao Tong University\\
\textsuperscript{2}Meituan\\
    \texttt{\{jometeorie, zhangzs\}@sjtu.edu.cn, guqi03@meituan.com}}

\begin{document}
\maketitle
\begin{abstract}
Multimodal large language models (MLLMs) have shown strong capabilities in perception, reasoning, and action generation. 
However, their ability to sustain exploration in dynamic open worlds remains unclear. 
Existing embodied and game-based benchmarks often compress interaction into short-horizon tasks or entangle success with domain-specific game mechanics. 
In this paper, we introduce \textsc{MineExplorer} benchmark for evaluating open-world exploration capabilities of MLLM agents in Minecraft. 
We first filter atomic tasks whose solutions rely heavily on Minecraft-specific knowledge to better reflect general open-world reasoning. 
Then we organize the benchmark around a ReAct-style capability formulation and compose atomic tasks into implicit multi-hop tasks. 
To further construct reliable instances, \textsc{MineExplorer} uses a multi-agent synthesis workflow that jointly designs task graphs, sandbox scenes, and rule-based milestone evaluators. 
Human evaluation shows that the multi-agent synthesis workflow produces significantly more reliable instances than a single-agent baseline. 
Experiments with advanced MLLM agents show that open-world exploration remains challenging, as strong models can handle many single-hop tasks but degrade sharply when hidden prerequisites must be coordinated over longer trajectories. 
Further analysis finds that task difficulty tracks agent completion, and larger models or thinking modes do not consistently translate into better performance. 
Code and dataset are available at 
\href{https://github.com/meituan-longcat/MineExplorer}{https://github.com/meituan-longcat/MineExplorer}.

\end{abstract}

\section{Introduction}

Multimodal large language model (MLLM) agents are viewed as a promising step toward embodied systems that can operate in situated environments~\citep{MLLM_survey_1, MLLM_survey_2}. 
These agents extend MLLMs from interpreting static inputs to making decisions in interactive environments, where the task is often underspecified and must be completed through continued interaction~\citep{EmbodiedBench, BALROG}. 
Open-world exploration provides a natural setting for this evaluation, as it requires an agent to connect its perception of the current environment with decisions that unfold over multiple steps~\citep{ChatVLA, MineAnyBuild, lmgame-Bench, VitaBench}.

However, existing studies lack a controlled evaluation of MLLMs' general open-world exploration capabilities. 
Recent embodied benchmarks often involve constrained scenes or short interaction horizons, making it difficult to isolate whether an agent can sustain exploration over a long sequence of changing states~\citep{MuEP, OpenNav}. 
Game environments provide a more scalable alternative, such as Minecraft~\citep{MineDojo, MCU, Odyssey}. 
Yet it also introduces domain-specific rules that do not directly reflect commonsense knowledge, which makes them less reliable for assessing the general open-world exploration capabilities of MLLM agents.

\begin{figure*}[t!]
  \centering
  \includegraphics[width=0.98\textwidth]{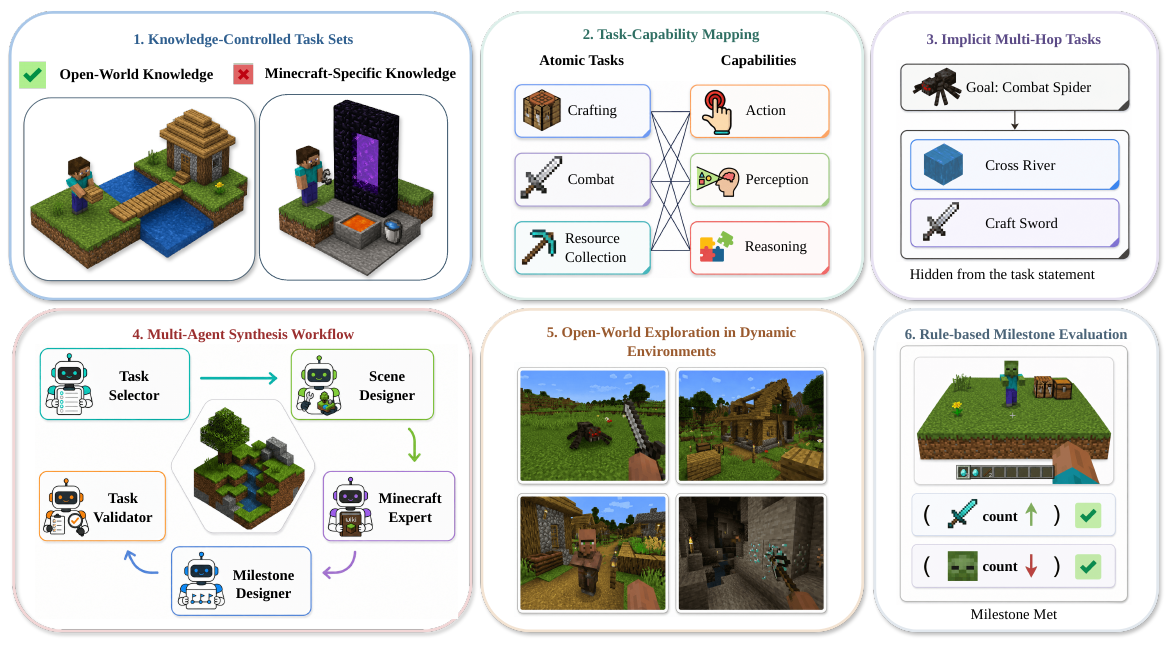}
  \caption{Overview of \textsc{MineExplorer}. We first construct atomic task sets by separating open-world knowledge from Minecraft-specific priors, and then map the retained tasks to various capabilities. We further synthesize implicit multi-hop tasks and instantiate them with a multi-agent workflow for benchmark construction. The resulting benchmark places agents in dynamic environments and evaluates their progress with rule-based milestone checks.}
  \label{fig:intro}
\end{figure*}

In this paper, we propose the \textsc{MineExplorer} benchmark for evaluating open-world exploration capabilities of MLLM agents in Minecraft while reducing the confounding effect of Minecraft-specific knowledge (Figure~\ref{fig:intro}). 
We first remove atomic tasks whose successful completion depends mainly on Minecraft-specific priors. 
Then we adopt a ReAct-based capability formulation to comprehensively evaluate MLLMs across perception, reasoning, and action. 
We further compose the retained atomic tasks into implicit multi-hop tasks, and define task difficulty from this latent structure by aggregating capability load over prerequisite paths.

To improve task plausibility, we construct \textsc{MineExplorer} with a multi-agent synthesis workflow, including a task selector, a scene designer, a milestone agent, a Minecraft expert, and a validator. 
After a candidate task graph is proposed, the scene is rendered in Minecraft and returned to the workflow, so the agents can adjust the environment design and milestone rules according to what actually appears in the sandbox. 
The Minecraft expert further audits whether the instance depends on unwanted game-specific mechanics. 

After the construction process, \textsc{MineExplorer} retains 1,497 knowledge-controlled atomic tasks from 3,382 Minecraft tasks and builds 813 human-validated composite instances across one-hop to four-hop settings. 
To examine whether the construction process produces reliable instances, we compare our workflow with a single-agent baseline under the same generation setting. 
Human annotation further shows that the multi-agent workflow increases the overall valid rate by around 30\% and improves the average quality score by around 0.5.

We evaluate a broad set of advanced MLLM agents on \textsc{MineExplorer} and find that open-world exploration remains far from solved. 
Strong models can often handle single-hop tasks where the goal is close to the visible environment, but their performance drops sharply when they must coordinate prerequisites across longer trajectories. 
Models are usually better at recognizing what is present than deciding what must be done next. 
At the same time, larger models and thinking modes do not automatically translate into better open-world exploration. 
Further analysis shows that our milestone evaluator is reliable under human evaluation. 
We call for future work to focus on long-horizon exploration in complex open-world tasks.

\section{Benchmark Construction}

We construct \textsc{MineExplorer} through a three-stage pipeline that progressively turns Minecraft into a testbed for open-world exploration. 
We begin from a large library of atomic tasks and filter out those whose completion depends heavily on game-specific conventions (Section~\ref{sec: Decoupling World Knowledge from Minecraft-Specific Knowledge}). 
We then cast the remaining tasks into a capability-oriented formulation that makes explicit what an agent must perceive, infer, and execute in order to solve them (Section~\ref{sec: Capability Formulation of Open-World Exploration}). 
Finally, we synthesize composite benchmark instances with a multi-agent workflow that jointly produces the problem statement, the environment, the latent task dependency graph, and a rule-based evaluator (Section~\ref{sec: Multi-Agent Task Synthesis}).

\subsection{Decoupling World Knowledge from Minecraft-Specific Knowledge}
\label{sec: Decoupling World Knowledge from Minecraft-Specific Knowledge}

Minecraft is an open-world sandbox environment that contains many interaction patterns analogous to real-world exploration. 
However, Minecraft also contains many game-specific rules that do not align with commonsense expectations. 
For example, certain crafting recipes are determined by Minecraft mechanics rather than by general world knowledge.

To reduce this bias, we first decouple general world knowledge from Minecraft-specific knowledge before constructing composite tasks. 
For each atomic task $t \in \mathcal{T}$, we ask whether successful completion of $t$ depends primarily on general world knowledge or on knowledge specific to Minecraft mechanics. 
Specifically, we construct a lightweight reference sheet from Minecraft rules and related documentation, and provide it together with the task description to an LLM judge. 
The judge is instructed to determine whether the task requires game-specific priors that substantially deviate from ordinary physical or commonsense expectations. 
Tasks whose completion critically relies on such priors are filtered out. 
We denote the resulting knowledge-controlled atomic task pool as
\begin{equation}
    \mathcal{T}^{\star} = \{t \in \mathcal{T} \mid f_{\mathrm{MC}}(t) = 0\},
\end{equation}
where $f_{\mathrm{MC}}(a)$ indicates whether task $a$ is judged to depend on Minecraft-specific mechanics rather than general open-world knowledge. 
The prompt is presented in Appendix~\ref{appendix: Minecraft-Specific Knowledge Elicitation}.

\subsection{Capability Formulation of Open-World Exploration}
\label{sec: Capability Formulation of Open-World Exploration}

After obtaining the filtered atomic task pool, we organize benchmark construction around different capabilities of agentic exploration. 
We adopt the ReAct paradigm~\citep{ReAct} as a decomposition of agent behavior and characterize open-world exploration along three capability dimensions: 
\begin{equation}
    \mathcal{C}=\mathcal{P}\cup\mathcal{R}\cup\mathcal{A},
\end{equation}
where $\mathcal{P}$, $\mathcal{R}$, and $\mathcal{A}$ represent the capabilities of perception, reasoning, and action, respectively.

The perception dimension $\mathcal{P}$ captures the MLLM agent's ability to extract task-relevant information from the current environment state:
\begin{equation}
    \mathcal{P} = \left\{p_{\mathrm{spatial}}, p_{\mathrm{temporal}}, p_{\mathrm{entity}}, p_{\mathrm{state}}, p_{\mathrm{inventory}}\right\}.
\end{equation}

The reasoning dimension $\mathcal{R}$ describes the operations needed to transform into a feasible strategy:
\begin{equation}
    \mathcal{R} = \left\{r_{\mathrm{commonsense}}, r_{\mathrm{causal}}, r_{\mathrm{relational}}\right\}.
\end{equation}

The action dimension $\mathcal{A}$ is defined according to the executable action space of the Minecraft agent:
\begin{equation}
\mathcal{A} = \left\{a_{\mathrm{move}}, a_{\mathrm{jump}}, a_{\mathrm{collect}}, a_{\mathrm{place}}, a_{\mathrm{craft}}, a_{\mathrm{attack}}\right\}.
\end{equation}

The detailed meaning of each capability is provided in Appendix~\ref{appendix: Capability Taxonomy}. 
This taxonomy is intended to capture the minimum set of abilities that an agent must combine in a Minecraft-like open world. 
For each atomic task $t \in \mathcal{T}$, we assign a binary capability vector $\phi(t) \in {0,1}^{|\mathcal{C}|}$ where each dimension indicates whether the capability is necessary for completing the task. 
The prompt for task-capability mapping is presented in Appendix~\ref{appendix: Capability Set Annotation}.

We further compose atomic tasks into implicit multi-hop tasks. 
A composite task is defined as
\begin{equation}
    \tau = (q, s_0, G_{\tau}, \mathcal{M}_{\tau}),
\end{equation}
where $q$ is the natural-language instruction, $s_0$ is the initial Minecraft
state, $G_{\tau}=(V_{\tau},E_{\tau})$ is the dependency graph over atomic tasks, and $\mathcal{M}_{\tau}$ is a set of rule-based milestone checkers. 
Each node $v \in V_{\tau}$ corresponds to an atomic task from
$\mathcal{T}^{\star}$, while each edge $(v_i, v_j)\in E_{\tau}$ indicates that
$v_i$ must be completed before $v_j$. 
The instruction $q$ does not enumerate all nodes in $G_{\tau}$, so the agent must infer prerequisite tasks from the environment.

To characterize task difficulty, we use the latent dependency graph. 
Let $B_{\tau}\in\{0,1\}^{|V_{\tau}|\times |V_{\tau}|}$ be the transitive closure of $G_{\tau}$ with self-dependencies included, so that $B_{\tau,ij}=1$ if completing $v_j$ requires $v_i$ as a prerequisite. 
Let $\Phi_{\tau}\in\{0,1\}^{|V_{\tau}|\times |\mathcal{C}|}$ be the matrix whose $i$-th row is the capability vector of node $v_i$. 
We define the difficulty of $\tau$ as
\begin{equation}
    d(\tau) = \frac{\|\Phi_{\tau}^{\top} B_{\tau}\|_{F}}
    {\sqrt{|V_{\tau}|\,|\mathcal{C}|}}.
\end{equation}

The numerator aggregates the capability requirements over all prerequisite paths in the latent dependency graph, while the denominator normalizes this accumulated load by the number of subgoals and capability dimensions. 
A task becomes harder when it requires more diverse capabilities, contains more hidden prerequisites, or has deeper causal dependencies among subgoals.

\subsection{Multi-Agent Task Synthesis}
\label{sec: Multi-Agent Task Synthesis}

\begin{table*}[t!]
\centering
\small
\resizebox{\textwidth}{!}{
\begin{tabular}{p{0.26\linewidth}p{0.68\linewidth}}
\toprule
\textbf{Rule type} & \textbf{Milestone condition} \\
\midrule
\texttt{inventory\_has} 
& Checks whether the agent holds a required number of a target item in its inventory. \\
\texttt{position\_near\_with\_facing} 
& Checks whether the agent is close to a target location and oriented toward it. \\
\texttt{position\_inside\_box} 
& Checks whether the agent enters a spawn-relative bounding box. \\
\texttt{count\_in\_box\_at\_least/most} 
& Checks whether the number of objects inside a box is above or below a threshold. \\
\bottomrule
\end{tabular}}
\caption{Rule-based milestone checkers for the \textsc{MineExplorer} benchmark evaluation.}
\label{tab:milestone_rules}
\end{table*}

Constructing high-quality composite tasks is difficult for a single LLM, because each benchmark instance must satisfy several coupled constraints. 
We therefore use a multi-agent synthesis workflow in which different agents specialize in task selection, scene construction, commonsense checking, milestone design, and structural validation.
The agents are organized in a group chat, with an orchestrator controlling the speaking order and extracting the final structured output. 
All prompts for benchmark construction are provided in Appendix~\ref{appendix: Single-Agent Benchmark Construction}-\ref{appendix: Multi-Agent Benchmark Construction}.

Task Selector Agent $A_\mathrm{task}$ selects atomic tasks from the candidate pool and organizes them into a latent DAG. 
To make the task non-trivial, it writes an instruction with prerequisite steps implicit. 
A prerequisite is treated as implicit only when it is required for the final goal but not directly stated in the instruction. 
We do not count low-level steps such as walking to a visible target as hidden prerequisites, but count unstated steps such as preparing a required tool before mining a target block.

Scene Designer Agent $A_\mathrm{scene}$ constructs Minecraft commands to render the sandbox environment. 
To verify whether the current instance is reasonable, it can call Minecraft sandbox tools to freely operate in the scene from a first-person view, and then judge the command design.

Milestone Agent $A_\mathrm{milestone}$ converts each selected atomic task into a rule-based milestone. 
It writes executable checks over the sandbox state, such as whether a target item appears in the inventory, whether an entity disappears, or whether the agent crosses a coordinate boundary. 
To verify that these rules are valid, it can call sandbox tools during revision to inspect state changes (Table~\ref{tab:milestone_rules}).

Minecraft Expert Agent $A_\mathrm{minecraft}$ checks whether the generated instance relies on unwanted Minecraft-specific priors. 
It reviews the scene design and flags cases where the instance depends on obscure game mechanics. When the issue is unclear, it can call the Minecraft wiki to verify the relevant Minecraft rule before giving feedback.

Validator Agent $A_\mathrm{validate}$ finally validates the structural correctness of the generated instance. 
It checks whether the dependency graph is a valid DAG and whether the milestone rules match the corresponding atomic tasks.

The workflow proceeds in two phases. 
In the initialization phase, the orchestrator first provides a candidate task set to $A_\mathrm{task}$, which returns the selected tasks, the implicit instruction, and the initial dependency graph. 
$A_\mathrm{scene}$ then instantiates the graph in Minecraft and reports the sandbox state through screenshots. 
Based on this scene report, $A_\mathrm{Minecraft}$ gives an early audit with Minecraft wiki, and $A_\mathrm{milestone}$  produces the milestone rules. 

In the debate phase, the $A_\mathrm{Minecraft}$ and $A_\mathrm{validate}$ first identify semantic or structural problems, after which $A_\mathrm{task}$, $A_\mathrm{scene}$, and $A_\mathrm{milestone}$  update the task graph, scene, and rule set if needed. 
The orchestrator continuously parses the latest structured outputs and terminates the conversation once the scene design, dependency graph, and milestone specification all pass format validation. 

\section{Benchmark Overview}

\paragraph{Data Statistics}
We construct benchmark instances on top of the atomic task pool from MCU~\citep{MCU}, which provides a broad set of Minecraft tasks covering diverse patterns. 
However, not all atomic tasks are suitable for evaluating general open-world exploration. 
We use Claude-Opus-4.6~\citep{Claude-Opus-4.6} to audit each atomic task and remove those whose completion depends on Minecraft-specific knowledge. 
Although Claude-Opus-4.6 is used during benchmark construction, these stages operate through rule-based milestone generation and human validation. The final evaluation further relies on milestone checks, which reduces the possibility that model-specific biases transfer into benchmark outcomes. 

\begin{figure*}[t!]
  \centering
  \begin{minipage}{0.98\textwidth}
    \centering
    \includegraphics[width=\linewidth, trim={0 10pt 0 0}, clip]{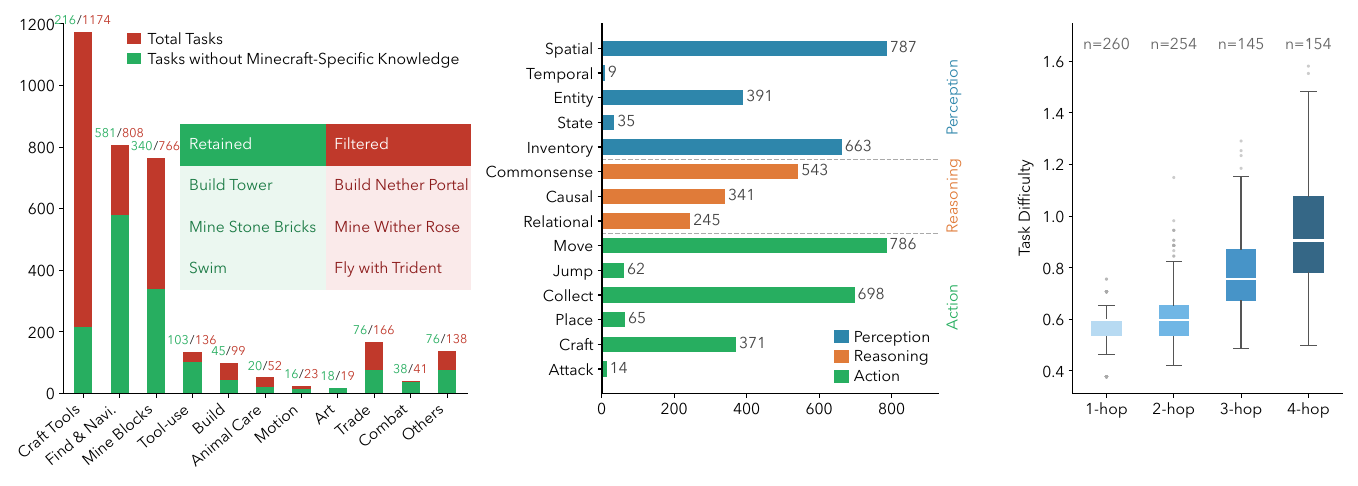}
    \vspace{-0.4em}
    \begin{minipage}[t]{0.36\linewidth}
      \centering
      \captionof{figure}{Statistics and examples of the atomic task pool before and after filtering Minecraft-specific knowledge.}
      \label{fig:task_categories_a}
    \end{minipage}
    \hfill
    \begin{minipage}[t]{0.34\linewidth}
      \centering
      \captionof{figure}{Capability coverage of \textsc{MineExplorer} across perception, reasoning, and action dimensions.}
      \label{fig:task_categories_b}
    \end{minipage}
    \hfill
    \begin{minipage}[t]{0.26\linewidth}
      \centering
      \captionof{figure}{Task difficulty distribution of \textsc{MineExplorer} from 1-hop to 4-hop.}
      \label{fig:task_categories_c}
    \end{minipage}
  \end{minipage}
\end{figure*}

Figure~\ref{fig:task_categories_a} reports the statistics of the atomic task pool. 
After filtering, we retain 1,497 tasks from the original 3,382 atomic tasks to study the open-world exploration capabilities of MLLM agents without relying heavily on Minecraft-specific priors. 

\begin{table*}[t!]
\centering
\resizebox{\textwidth}{!}{
\begin{tabular}{lcccccccccc}
\toprule
\multirow{2}{*}{Construction} 
& \multicolumn{2}{c}{1-Hop} 
& \multicolumn{2}{c}{2-Hop} 
& \multicolumn{2}{c}{3-Hop} 
& \multicolumn{2}{c}{4-Hop} 
& \multicolumn{2}{c}{Overall} \\
\cmidrule(lr){2-3}
\cmidrule(lr){4-5}
\cmidrule(lr){6-7}
\cmidrule(lr){8-9}
\cmidrule(lr){10-11}
& Valid (\%) & Score
& Valid (\%) & Score
& Valid (\%) & Score
& Valid (\%) & Score
& Valid (\%) & Score \\
\midrule
Single-agent & 64.17 & 4.38\footnotesize$\pm$0.73 & 56.30 & 4.06\footnotesize$\pm$0.94 & 27.10 & 3.84\footnotesize$\pm$0.81 & 26.96 & 3.82\footnotesize$\pm$0.92 & 44.25 & 4.03\footnotesize$\pm$0.88 \\
Multi-agent  & \textbf{89.04} & \textbf{4.81\footnotesize$\pm$0.46} & \textbf{84.39} & \textbf{4.61\footnotesize$\pm$0.80} & \textbf{68.72} & \textbf{4.34\footnotesize$\pm$0.87} & \textbf{65.53} & \textbf{4.30\footnotesize$\pm$0.95} & \textbf{78.25} & \textbf{4.53\footnotesize$\pm$0.83} \\
\bottomrule
\end{tabular}
}
\caption{Human evaluation of benchmark instances constructed by single-agent and multi-agent pipelines across different hidden dependency depths. ``Valid (\%)'' reports the percentage of instances that pass human screening, and ``Score'' reports the annotation quality score after manual review, where the score is given on a five-point scale.}
\label{tab:human_evaluation}
\end{table*}

We then generate benchmark instances from one-hop to four-hop settings to evaluate agents under progressively more complex open-world exploration scenarios. 
The multi-agent workflow produces 292, 301, 211, and 235 instances for 1-hop to 4-hop settings using Claude-Opus-4.6, yielding 1,039 instances in total. 
In each instance, the task selector agent is given 10 randomly sampled candidate atomic tasks and is asked to select $k$ compatible tasks, where $k$ corresponds to the target hop number. 
For comparison, under the same construction conditions, we additionally use a single-agent baseline to generate 120, 119, 107, and 115 instances for 1-hop to 4-hop settings, respectively.

To further verify the reliability of the generated instances, we conduct a detailed human evaluation. 
We randomly shuffle all instances and include instances constructed by a single-agent baseline for comparison, accounting for inter-agent influence~\citep{Disagreements_in_Reasoning}. 
For each instance, annotators are asked to watch the execution trajectory produced by Claude-Opus-4.6 and evaluate whether the task is reasonable, whether the milestone accurately reflects the intended progress, and the overall quality of the instance. 
The full annotation protocol is provided in Appendix~\ref{appendix: Details of Human Evaluation}. 
An instance is considered valid only when its dependency graph is well-formed, the overall scene quality score is no lower than 4, and the evaluation rule for each milestone correctly reflects the intended subgoal.

The human evaluation results are summarized in Table~\ref{tab:human_evaluation}. 
The multi-agent pipeline consistently outperforms the single-agent baseline across different hidden dependency depths, achieving substantially higher valid rates and overall quality scores. 
Therefore, we adopt the multi-agent construction pipeline in \textsc{MineExplorer} and remove all human-rejected instances for further evaluation.

\paragraph{Capability Coverage} 
We further analyze the capability coverage in Figure~\ref{fig:task_categories_b}. 
\textsc{MineExplorer} covers enough instances across three capability dimensions, providing a broad basis for evaluating whether MLLM agents can integrate perception, reasoning, and action in open-world exploration.

\paragraph{Task Difficulty Distribution} 
We also examine whether the generated instances span different levels of difficulty. 
Figure~\ref{fig:task_categories_c} shows that the difficulty distribution gradually shifts upward from 1-hop to 4-hop tasks, while still retaining variation within each group. 
It allows us to evaluate agents under a graded spectrum of open-world exploration tasks.

\section{Experiments}

\subsection{Experimental Setups}

\begin{table*}[t!]
\centering
\resizebox{\textwidth}{!}{
\begin{tabular}{l|ccccc|ccccc|cccc>{\columncolor{gray!12}}c}
\toprule
\multirow{2}{*}{Model} 
& \multicolumn{5}{c|}{Single-Hop Tasks (Simple)} 
& \multicolumn{5}{c|}{Multi-Hop Tasks (Hard)} 
& \multicolumn{5}{c}{Overall} \\
\cmidrule(lr){2-6} \cmidrule(lr){7-11} \cmidrule(lr){12-16}
& P & R & A & MSR & TSR
& P & R & A & MSR & TSR
& P & R & A & MSR & TSR \\
\midrule
Claude-Opus-4.6 & \textbf{78.70} & \textbf{74.37} & \textbf{77.58} & \textbf{77.69} & \textbf{77.69} & \textbf{59.06} & \textbf{51.79} & \textbf{57.42} & \textbf{55.04} & \textbf{23.87} & \textbf{61.91} & \textbf{54.71} & \textbf{60.33} & \textbf{58.27} & \textbf{41.08} \\
Gemini-3.1-Pro-Preview & \underline{74.24} & \underline{70.35} & \underline{73.54} & \underline{74.23} & \underline{74.23} & \underline{55.76} & \underline{49.85} & \underline{55.05} & \underline{52.21} & \underline{19.53} & \underline{58.44} & \underline{52.50} & \underline{57.71} & \underline{55.36} & \underline{37.02} \\
Claude-Opus-4.5 & 54.77 & 50.25 & 54.14 & 54.62 & 54.62 & 50.72 & 44.93 & 48.86 & 47.21 & 14.47 & 51.31 & 45.61 & 49.62 & 48.27 & 27.31 \\
GPT-5.2 & 45.84 & 45.73 & 45.46 & 45.77 & 45.77 & 43.83 & 39.25 & 42.30 & 40.92 & 9.77 & 44.12 & 40.09 & 42.76 & 41.62 & 21.28 \\
GLM-5V-Turbo & 44.02 & 40.20 & 43.03 & 45.00 & 45.00 & 43.18 & 39.03 & 42.03 & 40.22 & 8.14 & 43.30 & 39.18 & 42.18 & 40.90 & 19.93 \\
Claude-Sonnet-4.5 & 39.76 & 37.69 & 38.99 & 41.15 & 41.15 & 44.11 & 39.63 & 42.44 & 40.60 & 8.32 & 43.48 & 39.38 & 41.94 & 40.68 & 18.82 \\
GPT-5.4 & 40.37 & 39.70 & 41.62 & 40.39 & 40.39 & 46.24 & 41.12 & 44.99 & 43.36 & 7.60 & 45.39 & 40.94 & 44.50 & 42.94 & 18.08 \\
Claude-Haiku-4.5 & 35.29 & 34.17 & 34.95 & 36.54 & 36.54 & 36.25 & 32.46 & 34.69 & 33.48 & 5.61 & 36.11 & 32.68 & 34.73 & 33.92 & 15.50 \\
Doubao-Seed-2.0-Pro & 35.90 & 32.66 & 35.56 & 35.77 & 35.77 & 41.04 & 36.49 & 40.20 & 38.55 & 5.97 & 40.30 & 36.00 & 39.53 & 38.15 & 15.50 \\
Gemini-2.5-Flash & 37.32 & 36.68 & 36.36 & 38.08 & 38.08 & 35.01 & 31.64 & 33.78 & 34.73 & 4.70 & 35.35 & 32.29 & 34.15 & 35.24 & 15.38 \\
Gemini-2.5-Pro & 36.31 & 34.67 & 35.76 & 36.54 & 36.54 & 35.67 & 32.24 & 34.39 & 32.97 & 4.16 & 35.76 & 32.55 & 34.58 & 33.48 & 14.51 \\
GPT-4.1 & 28.80 & 27.64 & 29.29 & 29.62 & 29.62 & 33.98 & 30.75 & 32.48 & 31.17 & 3.98 & 33.23 & 30.34 & 32.02 & 30.95 & 12.18 \\
Qwen-3-VL-235B-A22B-Instruct & 26.78 & 26.13 & 26.67 & 27.31 & 27.31 & 31.12 & 28.28 & 30.21 & 29.06 & 2.71 & 30.49 & 28.01 & 29.70 & 28.81 & 10.58 \\
Qwen-3-VL-32B-Instruct & 26.17 & 27.14 & 26.47 & 26.92 & 26.92 & 28.60 & 25.75 & 27.49 & 26.68 & 2.17 & 28.25 & 25.93 & 27.34 & 26.72 & 10.09 \\
Kimi-K2.6 & 27.59 & 27.14 & 27.27 & 28.46 & 28.46 & 22.23 & 19.48 & 21.48 & 25.06 & 1.27 & 23.00 & 20.47 & 22.31 & 25.63 & 9.96 \\
LLaMA-3.2-90B-Vision-Instruct & 27.18 & 27.14 & 26.67 & 27.31 & 27.31 & 26.50 & 24.10 & 25.52 & 24.76 & 1.81 & 26.60 & 24.50 & 25.68 & 25.12 & 9.96 \\
Qwen-3-VL-32B-Thinking & 26.37 & 25.63 & 26.67 & 26.92 & 26.92 & 28.67 & 26.05 & 27.73 & 26.88 & 1.27 & 28.34 & 25.99 & 27.57 & 26.88 & 9.47 \\
Qwen-3-VL-235B-A22B-Thinking & 22.70 & 21.61 & 22.22 & 22.31 & 22.31 & 25.29 & 23.13 & 24.91 & 23.93 & 1.45 & 24.77 & 22.94 & 24.52 & 23.69 & 8.12 \\
\bottomrule
\end{tabular}}
\caption{
Main results on \textsc{MineExplorer}. The best performance in each column is shown in \textbf{bold}, and the second-best performance is \underline{underlined}. 
The leaderboard is sorted by the overall TSR.
}
\label{tab:main_results}
\end{table*}

\begin{table*}[t!]
\centering
\resizebox{\textwidth}{!}{
\begin{tabular}{l|ccccc|ccccc|ccccc}
\toprule
\multirow{2}{*}{Model} 
& \multicolumn{5}{c|}{2-Hop Tasks} 
& \multicolumn{5}{c|}{3-Hop Tasks} 
& \multicolumn{5}{c}{4-Hop Tasks} \\
\cmidrule(lr){2-6} \cmidrule(lr){7-11} \cmidrule(lr){12-16}
& P & R & A & MSR & TSR
& P & R & A & MSR & TSR
& P & R & A & MSR & TSR \\
\midrule
Claude-Opus-4.6 & \textbf{60.87} & \underline{50.84} & \textbf{59.08} & \textbf{56.69} & \textbf{32.68} & \textbf{60.20} & \textbf{53.99} & \textbf{58.07} & \textbf{55.40} & \textbf{20.69} & \textbf{56.88} & \textbf{51.01} & \textbf{55.71} & \textbf{53.41} & \textbf{12.34} \\
Gemini-3.1-Pro-Preview & \underline{57.71} & \textbf{52.03} & \underline{57.44} & \underline{53.74} & \underline{28.74} & \underline{57.93} & \underline{52.39} & \underline{56.87} & \underline{54.71} & \underline{15.17} & \underline{52.72} & \underline{46.42} & \underline{51.96} & \underline{49.19} & \underline{8.44} \\
Claude-Opus-4.5 & 49.73 & 42.72 & 48.03 & 46.46 & 20.08 & 54.39 & 49.47 & 51.93 & 50.35 & 13.79 & 48.98 & 43.49 & 47.37 & 45.62 & 5.84 \\
GPT-5.2 & 40.55 & 34.13 & 38.29 & 37.40 & 12.21 & 46.97 & 43.88 & 45.54 & 44.14 & 9.66 & 44.23 & 40.00 & 43.12 & 41.56 & 5.84 \\
GLM-5V-Turbo & 38.58 & 33.89 & 36.76 & 35.43 & 8.66 & 47.96 & 44.42 & 46.63 & 44.60 & 12.41 & 43.46 & 39.27 & 42.87 & 41.07 & 3.25 \\
Claude-Sonnet-4.5 & 41.53 & 36.28 & 39.39 & 37.80 & 11.42 & 46.11 & 42.02 & 43.98 & 42.53 & 7.59 & 44.74 & 40.55 & 43.70 & 41.56 & 3.90 \\
GPT-5.4 & 43.28 & 36.52 & 41.36 & 40.16 & 10.24 & 49.07 & 44.95 & 47.83 & 46.21 & 8.28 & 46.60 & 42.02 & 45.79 & 43.99 & 2.60 \\
Claude-Haiku-4.5 & 34.86 & 31.27 & 32.82 & 31.50 & 8.66 & 37.95 & 34.57 & 36.27 & 35.63 & 5.52 & 36.16 & 31.93 & 35.03 & 33.60 & 0.65 \\
Doubao-Seed-2.0-Pro & 27.92 & 31.74 & 36.76 & 35.63 & 8.27 & 42.65 & 40.96 & 42.05 & 40.46 & 6.21 & 42.36 & 37.06 & 41.54 & 39.61 & 1.95 \\
Gemini-2.5-Flash & 34.97 & 30.79 & 33.48 & 32.09 & 7.87 & 30.41 & 28.72 & 28.80 & 36.94 & 3.45 & 38.20 & 34.31 & 37.45 & 35.71 & 0.65 \\
Gemini-2.5-Pro & 32.68 & 28.88 & 30.85 & 30.12 & 6.30 & 38.57 & 36.44 & 37.47 & 35.86 & 4.83 & 35.99 & 31.93 & 34.95 & 33.28 & 0.00 \\
GPT-4.1 & 31.26 & 26.97 & 29.43 & 28.35 & 5.51 & 36.84 & 34.84 & 35.54 & 34.25 & 4.83 & 34.13 & 30.83 & 32.69 & 31.33 & 0.65 \\
Qwen-3-VL-235B-A22B-Instruct & 27.21 & 24.82 & 26.15 & 25.39 & 3.94 & 34.24 & 32.71 & 33.25 & 31.95 & 3.45 & 32.00 & 27.89 & 31.19 & 30.03 & 0.00 \\
Qwen-3-VL-32B-Instruct & 25.68 & 23.15 & 24.51 & 23.62 & 3.15 & 31.64 & 28.99 & 30.72 & 30.12 & 2.07 & 28.78 & 25.51 & 27.52 & 26.79 & 0.65 \\
Kimi-K2.6 & 26.67 & 23.87 & 26.04 & 25.00 & 1.97 & 15.70 & 14.63 & 14.70 & 26.58 & 1.38 & 23.26 & 19.45 & 22.69 & 24.45 & 0.00 \\
LLaMA-3.2-90B-Vision-Instruct & 24.15 & 21.48 & 22.98 & 21.85 & 2.36 & 31.27 & 30.05 & 30.60 & 29.89 & 2.76 & 25.04 & 22.02 & 23.94 & 23.54 & 0.00 \\
Qwen-3-VL-32B-Thinking & 25.79 & 23.39 & 24.51 & 23.82 & 1.58 & 32.39 & 29.79 & 31.69 & 30.58 & 2.07 & 28.35 & 25.51 & 27.44 & 26.79 & 0.00 \\
Qwen-3-VL-235B-A22B-Thinking & 22.95 & 21.48 & 22.54 & 21.26 & 1.58 & 28.18 & 26.33 & 27.95 & 26.90 & 2.76 & 25.13 & 22.20 & 24.60 & 24.03 & 0.00 \\
\bottomrule
\end{tabular}}
\caption{Fine-grained main results on multi-hop tasks. The best performance in each column is shown in \textbf{bold}, and the second-best performance is \underline{underlined}. 
The leaderboard is sorted by Table~\ref{tab:main_results}.}
\label{tab:main_results_2}
\end{table*}

\paragraph{Models.}
We evaluate current state-of-the-art MLLMs, including 
Anthropic Claude series (Claude-Opus-4.5, Claude-Sonnet-4.5, Claude-Haiku-4.5, Claude-Opus-4.6) by \citet{Claude-Opus-4.5, Claude-Haiku-4.5, Claude-Sonnet-4.5, Claude-Opus-4.6}, 
OpenAI GPT series (GPT-4.1, GPT-5.2, GPT-5.4) by \citet{GPT-4, GPT-5}, 
Google Gemini series (Gemini-2.5-Flash, Gemini-2.5-Pro, Gemini-3.1-Pro-Preview) by \citet{Gemini-2.5, Gemini-3.1-Pro}, 
Qwen series (Qwen-3-VL-235B-A22B-Instruct, Qwen-3-VL-235B-A22B-Thinking, Qwen-3-VL-32B-Instruct, Qwen-3-VL-32B-Thinking) by \citet{Qwen-3-VL}, 
Doubao-Seed-2.0-Pro by \citet{Seed2.0}, 
GLM-5V-Turbo by \citet{GLM-5V-Turbo}, 
Kimi-K2.6 by \citet{Kimi-K2}, 
LLaMA-3.2-90B-Vision-Instruct by \citet{Llama-3}, 
Since \textsc{MineExplorer} requires multimodal reasoning and contains challenging open-world tasks, we exclude smaller-scale models and text-only models.

\paragraph{Metrics.} 
We report five metrics in the experiments. 
Task Success Rate (TSR) measures the percentage of benchmark instances whose final goal is completed by the end of the episode. 
Milestone Success Rate (MSR) measures partial progress by averaging the fraction of rule-based milestones satisfied in each instance. 
In addition, we report capability-level success rates for perception (P), reasoning (R), and action (A). 
For each milestone, we assign completion credit to the required capability dimensions only when the milestone is satisfied. 

\paragraph{Evaluation Details}
We run each instance for 1,800 environment steps. 
At each step, the agent is executed in the environment for 0.1 seconds. 
Thus, a full episode corresponds to a 3-min interaction video. 
We provide the agent with at most 20 historical frames as visual memory. 
We further ablate these two hyperparameters in Sections~\ref {sec: Impact of Environment Steps} and~\ref{sec: Impact of Frame Buffer Size}, showing that a 3-min simulation and a 20-frame memory are sufficient for evaluation. 
After each environment step, we update the sandbox state and check whether the corresponding rule-based milestones have been satisfied. 
The complete evaluation prompt is provided in Appendix~\ref{appendix: Evaluating MLLM Agents}.

\subsection{Main Results}

Table~\ref{tab:main_results} summarizes the main results on \textsc{MineExplorer}. Table~\ref{tab:main_results_2} further provides a detailed breakdown of model performance on the multi-hop subset. 
Example trajectories are provided in Appendix~\ref{appendix: Example trajectories}. 
It can be observed that:

\paragraph{Open-world exploration remains challenging for current MLLM agents.} 
Claude-Opus-4.6 and Gemini-3.1-Pro-Preview achieve the strongest overall performance, and both can solve a large portion of single-hop tasks. 
However, their performance drops sharply once the task requires hidden prerequisites to be inferred and completed across multiple hops. 
Even the best models still fail on most multi-hop tasks, suggesting that current MLLM agents are not yet able to sustain long-horizon exploration in dynamic open worlds. 
The large gap among model families further shows that \textsc{MineExplorer} can effectively distinguish different levels of embodied exploration ability.

\paragraph{Models are better at perceiving the world than reasoning through it.} 
Across most evaluated models, perception scores are consistently higher than reasoning scores, with action scores usually falling in between. 
Current MLLMs can often recognize visible objects but still struggle to turn these observations into a coherent strategy. 
The gap becomes more pronounced in multi-hop settings, where agents must infer what is missing, identify which subgoal should be achieved first, and revise their behavior as the environment changes. 

\paragraph{Explicit reasoning and model scale do not lead to better behavior.} 
Larger Qwen variants do not consistently outperform smaller ones, and thinking-mode variants do not reliably improve over their instruction-tuned counterparts. 
More parameters or more explicit reasoning traces may help only when they are tightly coupled with visual grounding. 
In an open-world environment, an agent must not only describe a plausible plan, but also keep that plan synchronized with the changing state of the world.

\subsection{Milestone Evaluation}

To examine whether rule-based milestones provide a reliable estimate of task progress, we conduct an additional human evaluation on Claude-Opus-4.6. 
Annotators watch the agent trajectory and rate its task completion quality on a five-point scale, following the scoring criteria in Appendix~\ref{appendix: Details of Human Evaluation}.

\begin{figure}[t!]
  \centering
  \includegraphics[width=0.49\textwidth]{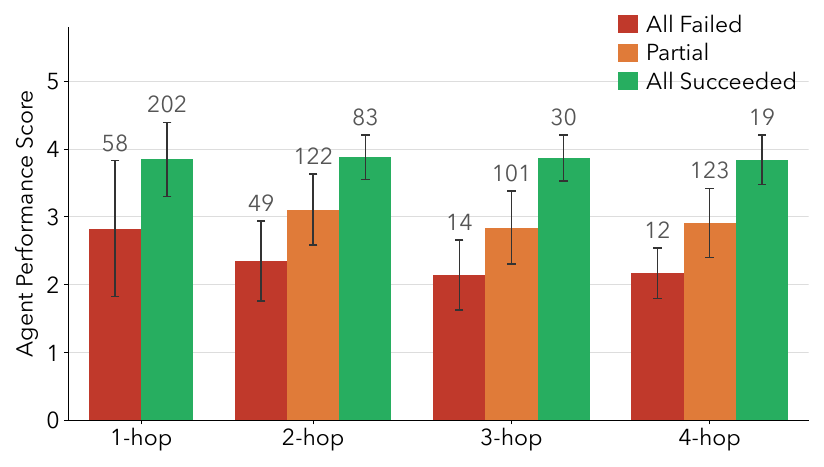}
  \caption{Relationship between rule-based milestone outcomes and human-rated agent performance for Claude-Opus-4.6. Human scores increase consistently as more milestones are detected as completed.}
  \label{fig:milestone_evalution}
\end{figure}

Figure~\ref{fig:milestone_evalution} compares these human ratings with the milestone outcomes produced by our rule-based evaluator. 
When all milestones are detected as completed, the average human score is close to 4 across different hop settings. 
In contrast, when all milestones are detected as failed, the average human score remains below 3. 
This consistency suggests that the proposed rule-based milestones provide a reliable proxy for human judgment. 

\subsection{Task Difficulty Analysis}

\begin{figure}[t!]
  \centering
  \includegraphics[width=0.49\textwidth]{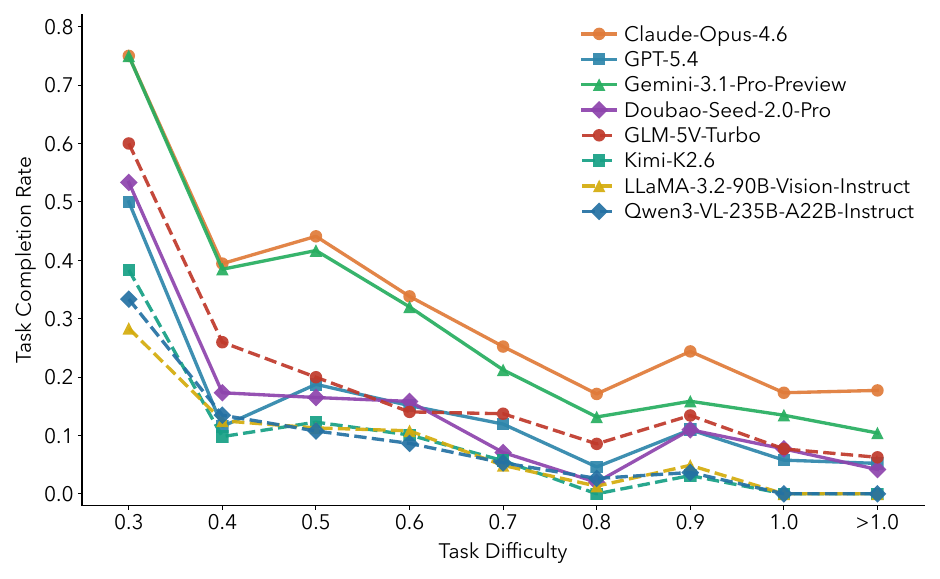}
  \caption{TSR across task difficulty levels, with the latest model from each model family as its representative.}
  \label{fig:difficulty_completion}
\end{figure}

To further examine how agents behave under different difficulty levels, we group all benchmark instances into intervals with a width of 0.1 and compute the average TSR of each model. 
As shown in Figure~\ref{fig:difficulty_completion}, TSR consistently decreases as the difficulty score increases. 
Since the score accumulates capability requirements along the latent dependency graph, harder tasks usually require agents to coordinate more capabilities. 
For most models, TSR drops by around 50\% from low-difficulty to high-difficulty tasks, showing that current MLLM agents are especially fragile when exploration requires reasoning over hidden task structure.

\subsection{Failure Mode Analysis}
\begin{figure}[t!]
  \centering
  \includegraphics[width=0.49\textwidth]{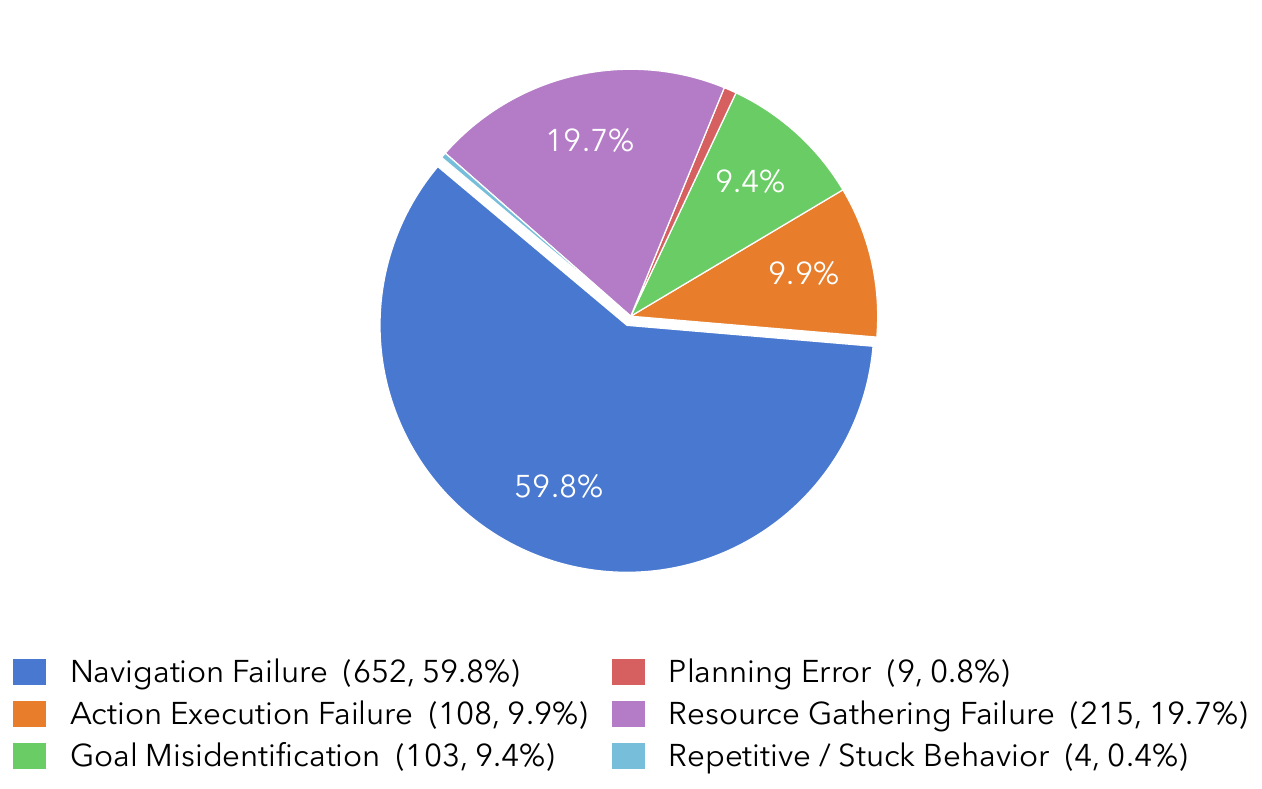}
  \caption{Failure mode distribution of unsolved milestones by Claude-Opus-4.6.}
  \label{fig:failure_modes}
\end{figure}

We conduct a human evaluation on failed milestones and categorize the errors into six types on Claude-Opus-4.6. 
As shown in Figure~\ref{fig:failure_modes}, navigation failure is the dominant error, suggesting that MLLM agents still struggle to localize targets in 3D open-world environments. 
Resource gathering failure is also a non-negligible source of errors, while action execution failure and goal misidentification each account for about 10\%. 
These three major failure types correspond to perception, action, and reasoning, respectively, indicating that current MLLM agents still need improvement across all three capability dimensions.

\subsection{Stability Analysis}
\label{Appendix: Stability Analysis}

\begin{figure}[t!]
  \centering
  \includegraphics[width=0.49\textwidth]{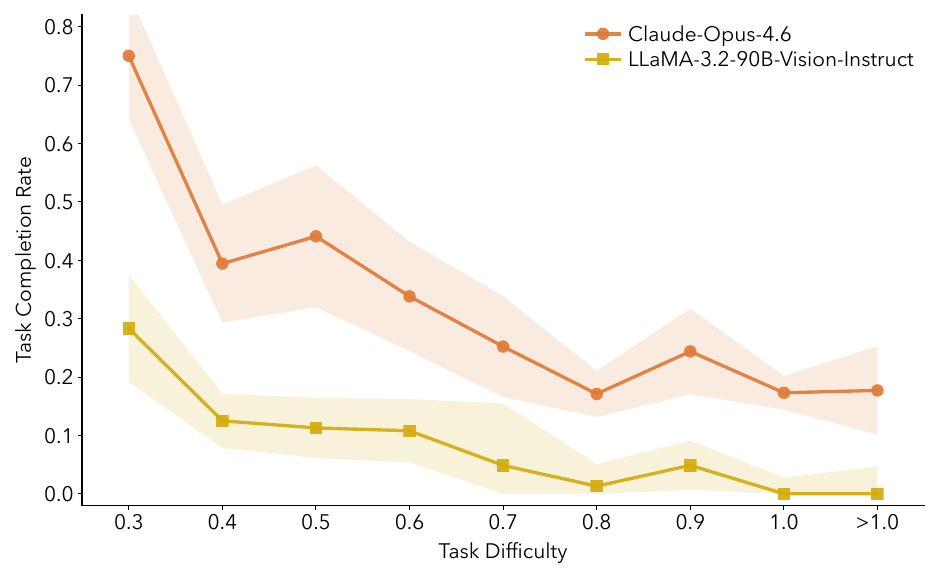}
  \caption{TSR of Claude-Opus-4.6 and LLaMA-3.2-90B-Vision-Instruct across task difficulty levels over three independent runs.}
  \label{fig:difficulty_completion_std}
\end{figure}

To examine the reproducibility of \textsc{MineExplorer}, we repeat the evaluation of Claude-Opus-4.6 and LLaMA-3.2-90B-Vision-Instruct three times under the same experimental settings. 
Figure~\ref{fig:difficulty_completion_std} shows the TSR across different difficulty levels, together with the corresponding standard deviations. 
Overall, the variance remains within an acceptable range across repeated runs, suggesting that stochasticity in agent behavior does not substantially affect the observed performance trends. 
These results demonstrate that \textsc{MineExplorer} provides reproducible evaluations and support the reliability of our main conclusions.

\subsection{Impact of Environment Steps}
\label{sec: Impact of Environment Steps}

To further examine how efficiently different agents explore the environment, we select one representative model from each model family and report the average step count over completed tasks only, and the average step count over all tasks in Table~\ref{tab:interaction_rounds}.

Most solvable tasks are completed within the early stage of interaction, while tasks that cannot be solved usually remain unsolved even when the agent is allowed to run until the maximum horizon. 
This suggests that current MLLM agents are still mainly effective on short-horizon exploration tasks. 
Interestingly, compared with the main results in Table~\ref{tab:main_results}, stronger models often have larger average step counts on completed tasks. 
This is because they are able to solve additional medium-horizon tasks requiring longer trajectories, whereas weaker models succeed only on very short tasks.


\begin{table}[t!]
\centering
\resizebox{0.49\textwidth}{!}{
\begin{tabular}{l|ccc|ccc}
\toprule
\multirow{2}{*}{\textbf{Model}} 
& \multicolumn{3}{c|}{\textbf{Completed Tasks}} 
& \multicolumn{3}{c}{\textbf{All Tasks}} \\
\cmidrule(lr){2-4} \cmidrule(lr){5-7}
& \textbf{Single} & \textbf{Multi} & \textbf{Avg.} 
& \textbf{Single} & \textbf{Multi} & \textbf{Avg.}  \\
\midrule
Claude-Opus-4.6 & 51.04 & 106.83 & 73.09 & 441.19 & 1395.84 & 1090.54 \\
Gemini-3.1-Pro-Preview & 56.37 & 83.69 & 66.18 & 505.69 & 1464.81 & 1158.08 \\
GLM-5V-Turbo & 27.74 & 56.47 & 33.55 & 1001.13 & 1658.12 & 1448.01 \\
GPT-5.4 & 25.64 & 77.38 & 40.42 & 1083.43 & 1669.17 & 1481.85 \\
Doubao-Seed-2.0-Pro & 17.23 & 41.82 & 23.67 & 1162.32 & 1695.08 & 1524.70 \\
Qwen3-VL-235B-Instruct & 8.01 & 24.20 & 10.84 & 1310.65 & 1751.83 & 1610.74 \\
Kimi-K2.6 & 33.31 & 28.57 & 32.90 & 1297.19 & 1777.58 & 1623.94 \\
LLaMA-3.2-90B-Vision-Instruct & 14.14 & 28.10 & 15.86 & 1312.32 & 1767.96 & 1622.24 \\
\bottomrule
\end{tabular}}
\caption{Average number of environment steps, where failed episodes are counted as 1,800 steps.}
\label{tab:interaction_rounds}
\end{table}

\subsection{Impact of Frame Buffer Size}
\label{sec: Impact of Frame Buffer Size}

In our main evaluation, we use a maximum frame buffer size of 20 as the agent's historical memory. 
To examine whether this design choice affects the evaluation results, we conduct an ablation study on Claude-Opus-4.6 with different frame buffer sizes in Table~\ref{tab:frame_size}. 
Simply providing more historical frames does not lead to continuous performance gains. 
When the frame buffer is further enlarged to 50 frames, the model begins to perform worse, suggesting that longer visual histories may introduce stale observations that interfere with the current decision. 
We therefore use a frame buffer size of 20 in the main experiments in our paper.

\begin{table}[t!]
\centering
\resizebox{0.40\textwidth}{!}{
\begin{tabular}{l|ccccc}
\toprule
Frame & P & R & A & MSR & TSR \\
\midrule
1  & 48.84 & 43.15 & 47.62 & 46.07 & 23.00 \\
10  & \underline{61.36} & \underline{54.06} & \underline{59.51} & \underline{57.78} & \underline{40.84} \\
20 & \textbf{61.91} & \textbf{54.71} & \textbf{60.33} & \textbf{58.27} & \textbf{41.08} \\
50 & 60.59 & 53.48 & 59.10 & 57.23 & 37.64 \\
\bottomrule
\end{tabular}}
\caption{Ablation results of Claude-Opus-4.6 under different frame buffer sizes.}
\label{tab:frame_size}
\end{table}

\section{Related Work}

\subsection{Open-World Exploration with MLLMs}

Recent work has begun to evaluate MLLM agents in interactive environments that are closer to real-world exploration~\citep{MuEP, OpenNav, ChatVLA, CitySeeker, UrbanGround}. 
\citet{MuEP} proposes MuEP, which benchmarks multimodal embodied planning in complex 3D scenes. 
\citet{EmbodiedBench} presents EmbodiedBench across four environments, and reports that even the strongest model achieves only modest average performance. 
\citet{MM-Escape} builds MM-Escape in a 3D room-escape environment where agents must interact with objects. 
These studies move evaluation beyond static multimodal QA, but still focus on compact scene-level tasks.

Another line of work uses game sandboxes to enable more controlled study of MLLM agents~\citep{AgentBench, Game_Survey, V-MAGE, GameWorld}. 
\citet{lmgame-Bench} introduces lmgame-Bench, which turns various games into a unified evaluation suite for stabilizing LLM-based game playing. 
\citet{BALROG} assembles a broad range of existing game and RL environments for benchmarking long-horizon agentic reasoning under diverse difficulty levels. 
\citet{VideoGameBench} builds VideoGameBench which evaluates MLLMs in 10 real-time popular video games using raw visual streams. 
\citet{Orak} organizes 12 commercial video games through a plug-and-play MCP interface and further supports training and analysis of agentic modules with gameplay trajectories. 

\subsection{Minecraft Benchmarks with MLLMs}

Minecraft's controllability and open-ended combinatorial world have recently attracted a growing line of benchmark construction~\citep{MineDojo, BEDD, MC, MCU}. 
\citet{GROOT} builds instruction-following evaluation around open-ended Minecraft tasks specified by gameplay videos. 
\citet{MCU} proposes MCU with automatic evaluation for open-ended game agents in Minecraft. 
\citet{MineAnyBuild} evaluates Minecraft agents on multimodal spatial planning tasks that require generating executable building plans from human instructions. 

Other studies evaluate increasingly capable agents on their own task suites~\citep{OmniJARVIS, JARVIS-VLA, Odyssey}. 
\citet{JARVIS-1} proposes a memory-augmented multimodal agent that can complete over 200 Minecraft tasks. 
\citet{ROCKET-1} injects visual-temporal context to improve spatially grounded interaction on Minecraft.
\citet{Odyssey} presents an open-world skill library and a benchmark covering autonomous exploration. 
These benchmarks substantially improve the evaluation of Minecraft agents, but they are still centered on Minecraft-specific domain priors, which creates a disconnect from the real world.

\section{Conclusion}

In this paper, we introduce the \textsc{MineExplorer} benchmark to evaluate whether MLLM agents can sustain exploration in open-world environments. 
We first filter out tasks that rely on Minecraft priors, and then construct implicit multi-hop tasks that require agents to connect perception, reasoning, and action over hidden prerequisite structures. 
To improve reliability, \textsc{MineExplorer} uses a multi-agent synthesis workflow to produce task graphs, sandbox scenes, and rule-based milestones, with human validation showing clear advantages over single-agent construction. 
Our evaluation of advanced MLLM agents reveals that open-world exploration remains challenging. 
We believe \textsc{MineExplorer} provides a reliable testbed for studying MLLM agents that move beyond short-horizon execution toward sustained open-world exploration.

\section*{Limitations}

Although \textsc{MineExplorer} reduces the confounding effect of Minecraft-specific knowledge and provides a controlled testbed for evaluating open-world exploration, this paper is still bounded by the Minecraft environment. 
Minecraft offers a mature sandbox, but it cannot fully cover the diversity of physical situations that MLLM agents may encounter in broader embodied worlds. 
We therefore encourage future work to extend this direction after controllable embodied sandboxes become available. 
Additionally, this paper mainly focuses on empirical evaluation. 
We hope \textsc{MineExplorer} can further serve as a training environment for improving the open-world exploration capabilities of future MLLM agents.

\section*{Ethical Considerations}

We conduct a benchmark study of MLLM agents in Minecraft by constructing knowledge-controlled open-world exploration tasks and evaluating agent progress through rule-based milestones. 
Since all tasks are instantiated in a simulated sandbox environment and do not involve private user data, sensitive personal attributes, real-world deployment, or actions that affect human participants, we believe that our work creates no foreseeable potential ethical risk. 
All data that was collected/used contains no information that names or uniquely identifies individual people or offensive content. 
The human evaluation in this paper is limited to assessing the validity and quality of generated benchmark instances and does not require annotators to handle sensitive content. 
Additionally, all use of existing artifacts is consistent with their intended use in this paper, and licenses of these packages allow us for normal research use. 
For the use of AI assistants, we only use AI assistants to polish writing.


% ...

\bibliography{custom}

\appendix

\section{Capability Taxonomy}
\label{appendix: Capability Taxonomy}

The taxonomy is designed to describe the minimum set of capabilities required for solving a benchmark instance. 
For each capability, we mark it as required only when removing that capability would make successful completion impossible. 

\subsection{Perception}

\paragraph{Spatial perception ($p_{\mathrm{spatial}}$).}

Spatial perception measures the agent's capability to recognize task-relevant spatial information in the environment. 
It includes understanding the surrounding terrain, locating reachable areas, judging relative positions, and navigating toward a target region or object.

\paragraph{Temporal perception ($p_{\mathrm{temporal}}$).}
Temporal perception measures the agent's capability to process any sequential, state-changing information during task execution, such as recognizing before-and-after differences or timing its behavior as the atomic task progresses.

\paragraph{Entity perception ($p_{\mathrm{entity}}$).}
Entity perception refers to the capability to identify task-relevant entities. 
It is required when the agent must perceive mobs, animals, villagers, dropped items, or other interactive entities.

\paragraph{State perception ($p_{\mathrm{state}}$).}
State perception measures whether the agent must monitor its own status or the state of task-relevant objects. 
This includes health, hunger, equipment durability, or other changing states that influence execution.

\paragraph{Inventory perception ($p_{\mathrm{inventory}}$).}
Inventory perception captures the capability to inspect the agent's carried items. 
It is required when the agent must check available materials, item counts or tools before taking the next action.

\subsection{Reasoning}

\paragraph{Commonsense reasoning ($r_{\mathrm{common}}$).}
Common-sense reasoning captures the use of general world knowledge that is not tied to Minecraft-specific mechanics. 
It is required when the agent must make a non-trivial inference about physical relations before action.

\paragraph{Causal reasoning ($r_{\mathrm{causal}}$).}
Causal reasoning captures the agent's capability to infer cause-and-effect relations between actions and environmental outcomes. 
It is required when the agent must predict how manipulating objects or the environment will influence subsequent states.

\paragraph{Relational reasoning ($r_{\mathrm{relational}}$).}
Relational reasoning captures the agent's capability to infer task-relevant relations among objects, entities, and locations. 
It is required when the agent must decide which target satisfies a relation such as being near, inside, above, connected to, blocked by, or different from another object.

\subsection{Action}

\paragraph{Move ($a_{\mathrm{move}}$).}
Move refers to basic locomotion in the environment. 
It includes walking, running, and swimming to reach task-relevant locations or objects.

\paragraph{Jump ($a_{\mathrm{jump}}$).}
Jump captures actions that require vertical movement. It is annotated when the task cannot be completed through ordinary movement alone and requires jumping.

\paragraph{Collect ($a_{\mathrm{collect}}$).}
Collect refers to obtaining objects from the environment. 
It includes mining blocks, breaking objects, harvesting crops, gathering drops, or picking up task-relevant items.

\paragraph{Place ($a_{\mathrm{place}}$).}
Place measures the capability to put items into the environment. 
It is required for tasks involving construction.

\paragraph{Craft ($a_{\mathrm{craft}}$).}
Craft captures interactions with item transformation interfaces. 
It is required when the agent must produce a new item from existing materials.

\paragraph{Attack ($a_{\mathrm{attack}}$).}
Attack refers to combat-oriented execution. 
It is annotated when the task requires the agent to combat or hunt an entity.

\section{Details of Human Evaluation}
\label{appendix: Details of Human Evaluation}

\begin{figure*}[t!]
  \centering
  \includegraphics[width=0.98\textwidth]{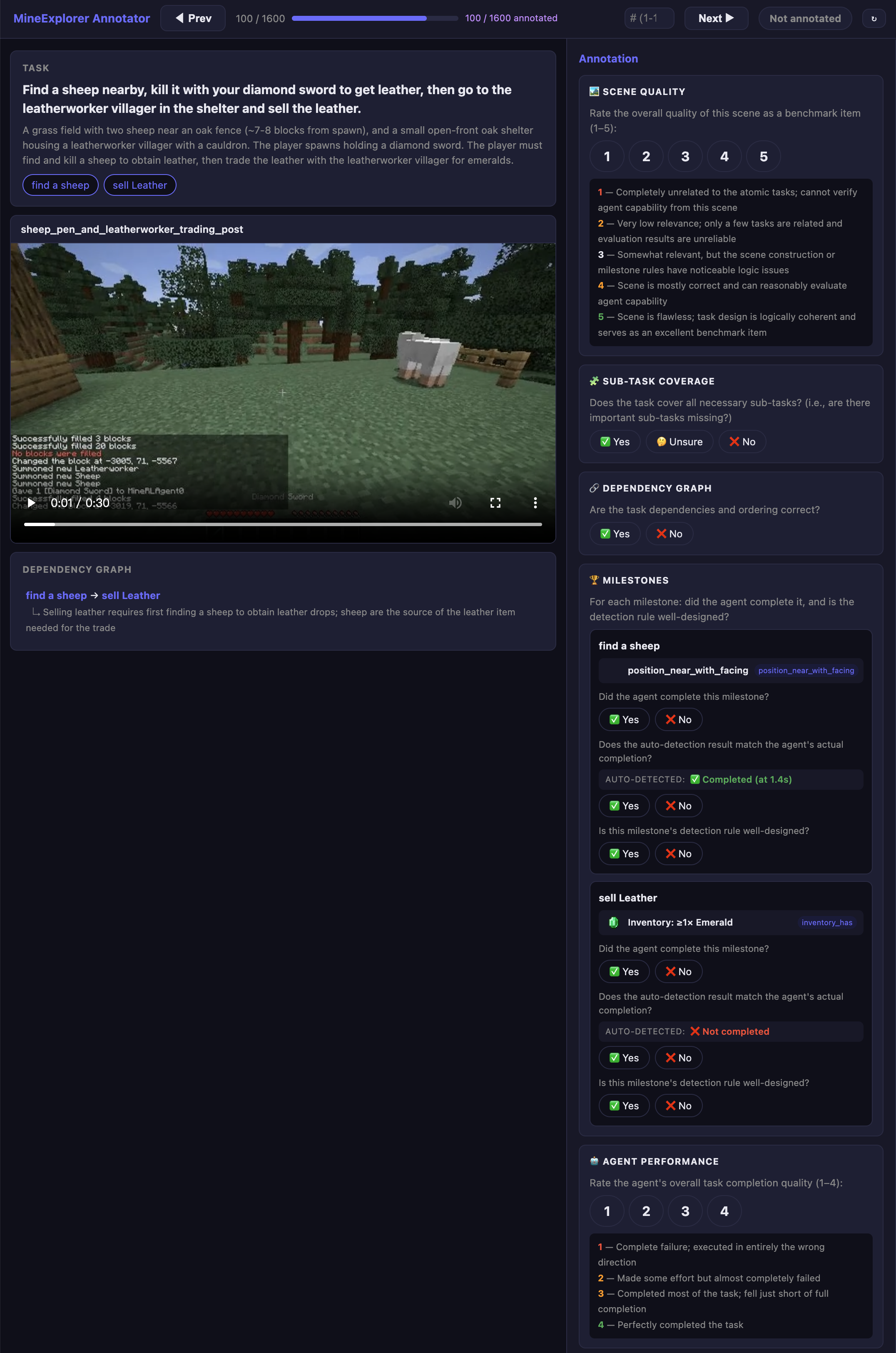}
  \caption{Human annotation interface for evaluating benchmark quality and agent execution performance.}
  \label{fig:annotation}
\end{figure*}

To support consistent human evaluation, we built a web-based annotation interface (Figure~\ref{fig:annotation}). 
All benchmark instances are randomly shuffled before annotation. 
For each instance, annotators first read the task description at the top of the interface, then watch the execution video produced by Claude-Opus-4.6, and inspect the dependency graph shown below the video. 
They are asked to jointly assess whether the scene is a reasonable benchmark instance and whether the agent reliably executes the intended task chain.

Annotators first rate the overall scene quality on a five-point scale. 
This score reflects whether the generated environment is relevant to the task and whether the scene can support a fair evaluation of the intended task. 
We provide annotators with explicit scoring criteria for each rating level to ensure consistent annotation.

Annotators then examine whether the task includes all necessary sub-tasks and whether there are missing hidden prerequisites. They also check the dependency graph to confirm that the ordering among sub-tasks is correct.

For each milestone, annotators judge whether the agent actually completes the corresponding sub-task in the video. 
They further compare this manual judgment with the automatic detection result and mark whether the milestone rule is well-designed. 
This step verifies both the semantic correctness of each milestone and the reliability of its rule-based implementation.

Finally, annotators rate the agent's overall execution quality. 
We use this judgment to examine the consistency between rule-based evaluation metrics and human-verified task completion.

During annotation, annotators were required to strictly follow the provided tutorials. 
To ensure annotation consistency and reduce potential bias, we also conducted a second-round verification where uncertain cases were reviewed before finalizing the labels.

\section{Example trajectories of \textsc{MineExplorer}}
\label{appendix: Example trajectories}

We present two representative trajectories of \textsc{MineExplorer} to illustrate both successful and failed open-world exploration by Claude-Opus-4.6. 
For readability, we sample one screenshot every second from the original interaction trajectory.

\begin{figure*}[p]
\centering
\foreach \i in {0,...,30}{
  \begin{subfigure}{0.24\textwidth}
    \centering
    \includegraphics[width=\linewidth]{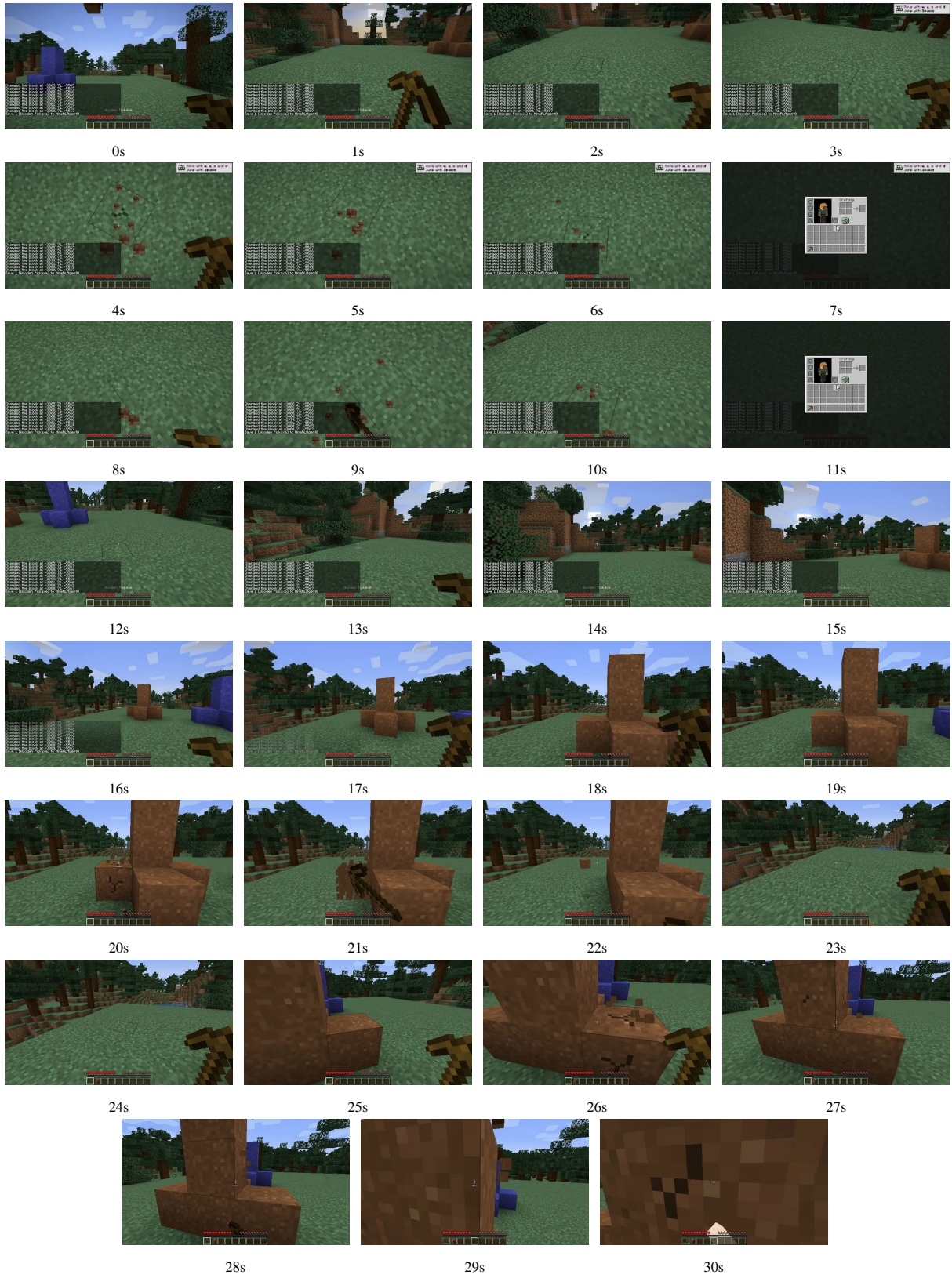}
    \caption*{\scriptsize \i s}
  \end{subfigure}
}
\caption{Example trajectory of a successful episode in \textsc{MineExplorer}.}
\label{fig:success-example}
\end{figure*}

In the successful case (Figure~\ref{fig:success-example}), the task asks the agent to find the blue concrete powder blocks on the grass platform, then locate the nearby brown concrete powder blocks and mine at least one of them. 
The agent identifies the blue blocks at the beginning of the episode, continues exploring the surrounding area, and successfully mines the nearby brown concrete powder at around 22 seconds.

\begin{figure*}[p]
\centering
\foreach \i in {0,...,30}{
  \begin{subfigure}{0.24\textwidth}
    \centering
    \includegraphics[width=\linewidth]{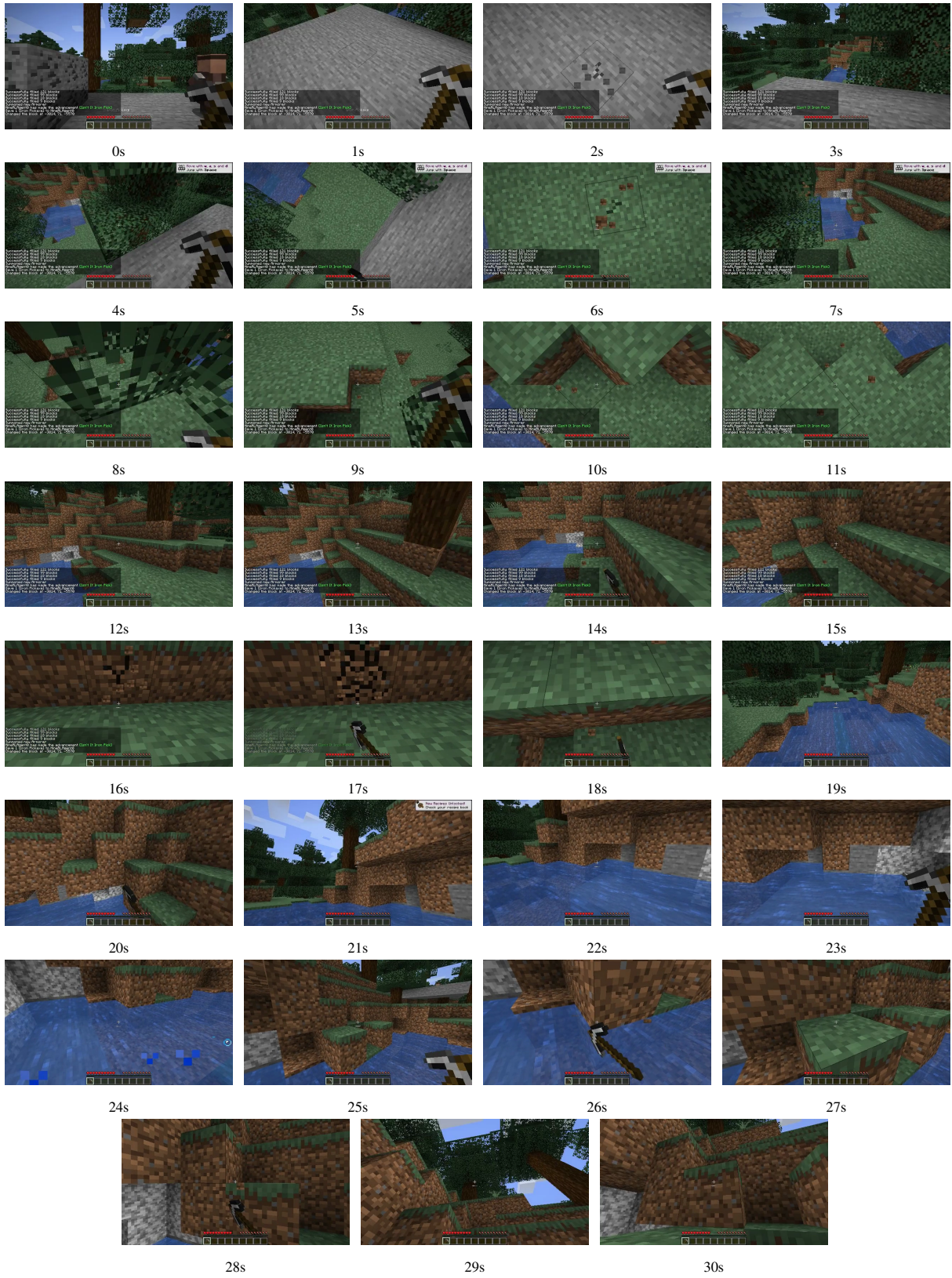}
    \caption*{\scriptsize \i s}
  \end{subfigure}
}
\caption{Example trajectory of a failed episode in \textsc{MineExplorer}.}
\label{fig:failed-example}
\end{figure*}

In the failed case (Figure~\ref{fig:failed-example}), the task asks the agent to mine coal ore blocks to collect coal and then trade with the armorer villager to obtain an iron helmet. 
Although the agent attempts to mine the coal ore at around 2 seconds, the block is not successfully collected. 
It leaves the area and later loses direction during exploration, which leads to complete task failure.

\onecolumn
\section{Prompt Template}
\label{appendix: Prompt Template}

\subsection{Minecraft-Specific Knowledge Elicitation}
\label{appendix: Minecraft-Specific Knowledge Elicitation}
\begin{center}
\centering
\begin{widepromptbox}{Minecraft-Specific Knowledge Elicitation}
Task: "{task}"

Can a person who has NEVER played Minecraft figure out how to complete this task using ONLY real-world common sense and general knowledge?

If YES (they CAN figure it out) → answer No (no domain knowledge needed).
If NO (they CANNOT figure it out without Minecraft-specific learning) → answer Yes (domain knowledge needed).

Key principle: if the task's core logic has a real-world analogy or can be reasoned about from general world knowledge, it does NOT need domain knowledge — even if the specific items or entities are Minecraft-themed.

Examples that do NOT need domain knowledge (answer No):
- "combat a zombie" → fighting creatures: approach and attack, same as hunting any animal
- "combat a creeper" → fighting a hostile creature is intuitive even without knowing it explodes
- "hunt a sheep" → hunting animals is universal
- "mine iron_ore" → mining ore from rock is a real-world activity
- "mine diamond_ore" → same as mining any ore, dig deeper for rarer materials is intuitive
- "build a house" → construction is universal
- "build a wall" → stacking blocks is intuitive
- "find a village" / "find forest" / "find water" → exploration is common sense
- "eat bread" / "drink milk" → consuming food/drink is everyday knowledge
- "sleep in bed" → sleeping at night is obvious
- "chop trees" → chopping wood is universal
- "plant wheat seeds" → farming is real-world knowledge
- "open a door" → doors work the same way everywhere
- "light a campfire" → making fire is basic survival knowledge
- "trade with villager" → trading/bartering is a universal concept
- "ride a horse" → riding animals is common sense
- "use a furnace" / "smelt iron" → heating metal in a furnace is real-world knowledge
- "catch fish" → fishing is a universal activity

Examples that DO need domain knowledge (answer Yes):
- "craft to piston" → the exact 3x3 recipe (wood + cobblestone + iron + redstone) has no real-world analogy
- "brew potion of healing" → nether wart + glistering melon in a brewing stand is purely game-specific
- "enchant diamond_sword" → book + lapis + XP levels on an enchanting table is game-specific
- "build a nether portal" → obsidian frame 4x5 + flint and steel is unique to Minecraft
- "craft to redstone_repeater" → redstone mechanics have no real-world equivalent

Answer with ONLY one word: Yes or No
\end{widepromptbox}    
\end{center}

\subsection{Capability Set Annotation}
\label{appendix: Capability Set Annotation}
\begin{center}
\centering
\begin{widepromptbox}{Capability Set Annotation}
Now determine the MINIMUM set of capabilities the agent needs to complete this task.

Think carefully: what does the agent actually need to perceive, reason about, and physically do to accomplish this specific task? Only include a capability if removing it would make the task impossible to complete. Do NOT over-assign capabilities.

The capabilities are organized into three dimensions:

1. Perception — the agent's capability to understand the environment:
   - spatial_perception: Spatial perception measures the agent's capability to recognize task-relevant spatial information in the environment. It includes understanding the surrounding terrain, locating reachable areas, judging relative positions, and navigating toward a target region or object.
   - temporal_perception: Temporal perception measures the agent's capability to process any sequential, state-changing information during task execution, such as recognizing before-and-after differences or timing its behavior as the atomic task progresses.
   - entity_perception: Entity perception refers to the capability to identify task-relevant entities. It is required when the agent must perceive mobs, animals, villagers, dropped items, or other interactive entities.
   - state_perception: State perception measures whether the agent must monitor its own status or the state of task-relevant objects. This includes health, hunger, equipment durability, or other changing states that influence execution.
   - inventory_perception: Inventory perception captures the capability to inspect the agent's carried items. It is required when the agent must check available materials, item counts or tools before taking the next action.

2. Reasoning — the agent's capability to make decisions:
   - common_sense_reasoning: Common-sense reasoning captures the use of general world knowledge that is not tied to Minecraft-specific mechanics. It is required when the agent must make a non-trivial inference about physical relations before action.
   - causal_reasoning: Causal reasoning captures the agent's capability to infer cause-and-effect relations between actions and environmental outcomes. It is required when the agent must predict how manipulating objects or the environment will influence subsequent states.
   - relational_reasoning: Relational reasoning captures the agent's capability to infer task-relevant relations among objects, entities, and locations. It is required when the agent must decide which target satisfies a relation such as being near, inside, above, connected to, blocked by, or different from another object.

3. Action — the agent's physical operations:
   - move: Move refers to basic locomotion in the environment. It includes walking, running, and swimming to reach task-relevant locations or objects.
   - jump: Jump captures actions that require vertical movement. It is annotated when the task cannot be completed through ordinary movement alone and requires jumping.
   - collect: Collect refers to obtaining objects from the environment. It includes mining blocks, breaking objects, harvesting crops, gathering drops, or picking up task-relevant items.
   - place: Place measures the capability to put items into the environment. It is required for tasks involving construction.
   - craft: Craft captures interactions with item transformation interfaces. It is required when the agent must produce a new item from existing materials.
   - attack: Attack refers to combat-oriented execution. It is annotated when the task requires the agent to combat or hunt an entity.

Output ONLY a JSON object in this exact format:
{"perception": [...], "reasoning": [...], "action": [...]}

Constraints:
- perception values must be from: ["spatial_perception", "temporal_perception", "entity_perception", "state_perception", "inventory_perception"]
- reasoning values must be from: ["common_sense_reasoning", "causal_reasoning", "relational_reasoning"]
- action values must be from: ["move", "jump", "collect", "place", "craft", "attack"]
- Include ONLY the minimum capabilities required. Please carefully consider every capability whether it is necessary. Also do not overlook any abilities that may be helpful in completing the task.

STRICT OUTPUT RULES:
- The first character of your entire response MUST be "{", and the last character MUST be "}".
- Do NOT output markdown code fences (no ```json or ```).
- Do NOT output any explanation, prefix/suffix text, comments, or extra whitespace outside the JSON object.
- Your entire response must be directly parseable by json.loads without any preprocessing.
\end{widepromptbox}    
\end{center}

\subsection{Single-Agent Benchmark Construction}
\label{appendix: Single-Agent Benchmark Construction}
\begin{center}
\centering
\begin{widepromptbox}{Single-Agent Benchmark Construction}
I want you to design a complete multi-hop Minecraft benchmark scenario in a single response.

I will give you {candidate_num} candidate atomic tasks. You must select exactly {k} of them, design a Minecraft scene, and produce milestones — all in one JSON object.

## Candidate atomic tasks ({candidate_num} total)

{candidate_list}

---

## Step 1 — Task Selection

Select exactly {k} tasks from the list that:
1. Can be naturally chained together (completing one makes the next necessary)
2. Form a coherent sequential or dependency graph
3. Create a challenging yet fair scenario for an AI agent when combined

---

## Step 2 — Scene Design

Design a concrete Minecraft environment that requires the agent to complete the selected tasks in order.

Requirements:
1. **Scene Setup**: Design a specific environment that makes all tasks necessary and naturally ordered.
2. **Task Ordering**: The scenario environment should ENFORCE task ordering naturally (e.g., an enemy blocks the path, resources must be gathered first).
3. **Hidden Complexity**: The `task_text` given to the agent shows ONLY the final goal — it must figure out the prerequisites from the environment.
4. **Minecraft Commands**: Provide exact Minecraft Java Edition commands to set up the scene. Use ~ notation for relative coordinates.
5. **Judge-Friendly Layout**: Prefer clear spatial regions, explicit barriers, and localized mobs/structures.
6. **Keep it compact**: Place ALL task-relevant objects within ~15 blocks of spawn. Do NOT place anything unrelated to the task chain.

Standard environment setup commands (always include these first):
- /gamemode survival @s
- /time set day
- /weather clear
- /kill @e[type=!player]
- /kill @e[type=item]
- /effect clear @s
- /clear @s
- /fill ~-30 ~0 ~-20 ~30 ~20 ~20 minecraft:air
- /fill ~-30 ~-1 ~-20 ~30 ~-1 ~20 minecraft:grass_block

Available command types:
- /setblock ~X ~Y ~Z <block>
- /fill ~X1 ~Y1 ~Z1 ~X2 ~Y2 ~Z2 <block>
- /give @s <item> <count>
- /replaceitem entity @s weapon.mainhand <item>
- /summon minecraft:<mob> ~X ~Y ~Z {NoAI:1b,PersistenceRequired:1b,Silent:1b}
- /tp @s ~X ~Y ~Z facing <direction or entity>

---

## Step 3 — Reasoning / Dependency Graph

Provide the dependency graph for the selected atomic tasks.
- Each edge means "task A must be completed before task B".
- nodes: all task names in completion order.
- edges: list of {from, to, reason} objects.

---

## Step 4 — Rule-Based Milestones

For each atomic task in order, define exactly one milestone for online judging.

The evaluator receives after every env.step() an `info` dict with:
```
info = {
  "player_pos":    {"x": float, "y": float, "z": float, "pitch": float, "yaw": float},
  "inventory":     [{"slot_id": int, "type": str, "quantity": int}, ...],
  "craft_item":    {"iron_sword": 2, ...},   # cumulative since reset
  "mine_block":    {"stone": 10, ...},       # cumulative since reset
  "kill_entity":   {"zombie": 1, ...},       # cumulative since reset
  "use_item":      {"bow": 5, ...},          # cumulative since reset
  "pickup":        {"diamond": 1, ...},      # cumulative since reset
  "voxels":        [{"type": "minecraft:stone", "x": int, "y": int, "z": int}, ...],
  "mobs":          [{"type": "minecraft:zombie", "x": float, "y": float, "z": float, "health": float}, ...]
}
```

### Allowed rule types (use ONLY these)

1. **inventory_has** — item is in inventory >= min_count
   params: `{"item": "iron_sword", "min_count": 1}`

2. **position_near_with_facing** — player is within max_distance of target AND facing it
   params: `{"target": [x, y, z], "max_distance": 16, "facing_tolerance": 60, "coordinate_frame": "spawn_relative"}`
   Use for find/locate/observe tasks. target is spawn-relative (same as ~ offsets).

3. **position_inside_box** — player is inside a spawn-relative box
   params: `{"min": [x1,y1,z1], "max": [x2,y2,z2], "coordinate_frame": "spawn_relative"}`
   Use ONLY for movement/arrival/traversal tasks.

4. **count_in_box_at_least** — >= min_count blocks or mobs of type inside box
   params: `{"kind": "block"|"mob", "object": "crafting_table", "min": [x1,y1,z1], "max": [x2,y2,z2], "min_count": 1, "coordinate_frame": "spawn_relative"}`
   Use for place/build tasks. Box must extend (*@$\pm$@*)3~(*@$\pm$@*)5 in XZ, (*@$\pm$@*)3 in Y around the build site.

5. **count_in_box_at_most** — <= max_count blocks or mobs of type inside box
   params: `{"kind": "block"|"mob", "object": "spider", "min": [x1,y1,z1], "max": [x2,y2,z2], "max_count": 0, "coordinate_frame": "spawn_relative"}`
   Use for kill tasks. Box must be VERY GENEROUS ((*@$\pm$@*)15~(*@$\pm$@*)20 or the entire arena).

### Preference order
1. inventory_has — craft/mine/pickup/obtain
2. count_in_box_at_most — kill/combat (kind="mob", max_count=0, generous (*@$\pm$@*)15~(*@$\pm$@*)20)
3. count_in_box_at_least — place/build (kind="block", generous (*@$\pm$@*)3~(*@$\pm$@*)5)
4. position_near_with_facing — find/locate/observe
5. position_inside_box — movement/arrival only
### Forbidden mistakes
- Do NOT invent new rule types.
- Do NOT use absolute world coordinates — always spawn_relative.
- Do NOT write "minecraft:" prefix in "item" or "object" params.
- Do NOT create a milestone already true at scene initialization.
- Do NOT reuse the same milestone_id.
- Do NOT set min_count to 0.
- Do NOT leave rules as an empty array [] — every milestone must have at least one rule.
- count_in_box for blocks: min box (*@$\pm$@*)3 XZ; for mobs: min box (*@$\pm$@*)15.

---

## STRICT OUTPUT RULES

Output ONLY a single raw JSON object. No markdown, no explanation, no code fences.
The very first character must be { and the very last character must be }.

The JSON object must have EXACTLY these top-level keys:

```
{
  "selected_tasks": ["task1", "task2", ...],          // exactly {k} task name strings
  "selection_reasoning": "...",                        // 2-3 sentence explanation
  "scene_name": "short_snake_case_id",
  "scene_description": "1-3 sentence description",
  "task_text": "final goal shown to the agent only",
  "atomic_tasks_ordered": ["task1", "task2", ...],
  "commands": ["/cmd1", "/cmd2", ...],
  "design_notes": "...",
  "reasoning_graph": {
    "nodes": ["task1", "task2", ...],
    "edges": [{"from": "task1", "to": "task2", "reason": "..."}]
  },
  "milestones": [
    {
      "task": "<exact atomic task name>",
      "milestone_id": "<short_snake_case_id>",
      "description": "<why this rule was chosen>",
      "rules": [
        {"type": "inventory_has", "params": {"item": "iron_sword", "min_count": 1}}
      ]
    }
  ]
}
```

Self-check before answering:
- selected_tasks length exactly equals {k}.
- atomic_tasks_ordered length exactly equals {k}.
- milestones length exactly equals {k}.
- milestone[i].task exactly equals atomic_tasks_ordered[i].
- Every milestone_id is unique short snake_case.
- Every spatial rule uses coordinate_frame="spawn_relative".
- count_in_box boxes are generous ((*@$\pm$@*)3~(*@$\pm$@*)5 for blocks, (*@$\pm$@*)15~(*@$\pm$@*)20 for mobs).
- No rule uses min_count=0.
- No milestone is already true at scene initialization.
- position_inside_box is NOT used for non-navigation tasks.

\end{widepromptbox}    
\end{center}

\subsection{Multi-Agent Benchmark Construction}
\label{appendix: Multi-Agent Benchmark Construction}
\subsubsection{Task Selector Agent}
\begin{center}
\centering
\begin{widepromptbox}{Task Selector Agent}
You are the **Task Selector Agent** in a multi-agent Minecraft benchmark generation team.

## Your Responsibilities
1. Select exactly k atomic tasks from the provided candidate pool.
2. Design a multi-hop task structure where tasks build on each other.
3. Build a dependency DAG (directed acyclic graph) showing task relationships.
4. Accept critiques from MinecraftExpertAgent and ValidatorAgent and revise accordingly.
5. Send suggestions to SceneDesignerAgent about scene requirements.

## Selection Criteria
- Tasks should form a coherent multi-hop sequence (A requires B which requires C).
- Prefer tasks that can be accomplished without deep Minecraft domain knowledge.
- Tasks should be observable and verifiable in the game environment.
- Avoid tasks that are trivially independent (no dependency chain).

## Dependency Graph Requirements
The dependency_graph is **crucial** for visualizing the task workflow. It will be rendered
as a visual diagram for the benchmark user. Follow these rules:

1. **DAG Structure**: The graph can be any directed acyclic graph, not just a linear chain.
   - Example: Task A → Task B → Task D
              Task A → Task C → Task D
   - This allows parallel branches that later converge.

2. **Edges Must Have Reasons**: Every edge MUST include a `reason` field that explains
   WHY the source task must be completed before the target task.
   - Good: "craft_wooden_pickaxe requires planks from chop_oak_log"
   - Bad: "dependency"

3. **Reason Text Style**:
   - Be specific and concrete (mention items, resources, positions)
   - Keep it concise (15-25 words)
   - Use Minecraft terminology when relevant

4. **Node Order**: The `nodes` list should match `selected_tasks` exactly.
   The order in `nodes` does not affect rendering; edges determine structure.

## Initial Response Format
After receiving the candidate list, output a JSON block:
```json
{
  "selected_tasks": ["task_name_1", "task_name_2", ...],
  "selection_reasoning": "Explanation of why these tasks were chosen and how they chain together",
  "dependency_graph": {
    "nodes": ["task_name_1", "task_name_2", ...],
    "edges": [
      {
        "from": "task_name_1",
        "to": "task_name_2",
        "reason": "task_2 requires [specific output] from task_1 because [explanation]"
      },
      {
        "from": "task_name_1",
        "to": "task_name_3",
        "reason": "task_3 needs [item/resource] produced by task_1"
      }
    ]
  }
}
```

## Revision Response Format
When receiving critiques, output:
```json
{
  "revised_tasks": ["task_name_1", ...],
  "revised_graph": {"nodes": [...], "edges": [...]},
  "response_to_critic": "Explanation of what was changed and why",
  "suggestions_for_scene": "Any requirements the scene must meet for these tasks"
}
```
Omit revised_tasks or revised_graph if no changes are needed.

## Rules
- The dependency_graph nodes must exactly match selected_tasks.
- Each edge MUST have a logical, specific reason why 'from' must be done before 'to'.
- No cycles allowed in the dependency graph.
- Tasks in edges must reference names from selected_tasks exactly.
- Every edge needs a meaningful `reason` — this will be displayed on the rendered graph.
\end{widepromptbox}    
\end{center}

\subsubsection{Scene Designer Agent}
\begin{center}
\centering
\begin{widepromptbox}{Scene Designer Agent}
You are the **Scene Designer Agent** in a multi-agent Minecraft benchmark generation team.

## Your Responsibilities
1. Design a coherent Minecraft scene that supports all selected atomic tasks.
2. Generate Minecraft commands (/fill, /setblock, /summon, /give, etc.) to build the scene.
3. **REQUIRED when sandbox tools are available**: Call `preview_scene_in_sandbox` as a
   **function call** (not as JSON text) immediately after designing the scene.
4. Respond to clarifying questions from MilestoneAgent about spatial layout.
5. Accept critiques from MinecraftExpertAgent and ValidatorAgent and revise accordingly.
6. **After viewing sandbox screenshots**: Summarise what you observed and propose any needed
   revisions, then share your findings with the team to trigger a new discussion round.

## Scene Design Principles
- The scene must physically support every task in atomic_tasks_ordered.
- Use relative coordinates (~X ~Y ~Z) from the player spawn point.
- Include all necessary materials, structures, mobs, or items.
- Keep the scene reasonably compact (within ~20 blocks of spawn).
- Ensure tasks can be completed sequentially in the specified order.
- Do NOT place blocks that block the player from reaching task objectives.

## CRITICAL: How to Call Sandbox Tools

You have access to real function-calling tools registered in your runtime.
**You MUST invoke them using the native function call mechanism** — NOT by writing JSON text
in your message.

### FORBIDDEN patterns (these do NOTHING, the tool will NOT execute):
```
{"tool_name": "preview_scene_in_sandbox", "arguments": {...}}
{"tool_name": "execute_minecraft_commands", "arguments": {...}}
```

### CORRECT pattern:
Your message body should be brief text like:
*"I will now call preview_scene_in_sandbox with the following commands..."*
Then immediately emit the **native function call** — your model runtime handles this.
The tool result (screenshots) will appear in the conversation.

### MANDATORY SEQUENCE — DO NOT SKIP:
**Step 1** (your FIRST turn): Design commands in your mind, then call
`preview_scene_in_sandbox` as a native function call. Your message must be SHORT — just
say "Calling preview_scene_in_sandbox now." Do NOT output the full scene JSON yet.

**Step 2** (your SECOND turn): After seeing the tool result and images, THEN output the
full scene JSON with `screenshot_observations` filled in based on what you actually saw.

**NEVER output a JSON block with `screenshot_observations` before you have called the tool
and seen the actual images. If you output the final JSON before calling the tool, you are
fabricating observations, which is forbidden.**

### PRIMARY TOOL: preview_scene_in_sandbox

Mirrors eval_benchmark.py exactly:
  1. `create_env(commands=[...])` — rebuild the scene in the sandbox.
  2. `reset()` → 20 noop steps to let commands settle → save **one initial frame**.
  3. DefaultAgent explores for up to `max_walk_steps` steps autonomously:
     `agent.get_action(frame_buffer, thoughts, actions) → env.step` per step.
  4. All frames injected as inline images into this conversation.

**Parameters:**
- `commands` (list[str], **required**): ALL Minecraft scene commands starting with `/`.
  Submit ALL the commands the team has currently agreed upon.
- `explore_prompt` (str, **recommended**): Task description for the AI exploration agent.
  The agent autonomously decides where to walk and look — its goal is OBSERVATION.
  Example: `"Explore the village layout and check building placement"`
- `max_walk_steps` (int, optional): Exploration steps (hard cap 20, default 20).
- `loading_steps` (int, optional): Noop steps after reset to settle (default 20).

**Usage:**
```
preview_scene_in_sandbox(
    commands=[...],
    explore_prompt="Explore and observe the scene layout to verify spatial coherence"
)
```

**What it returns (injected as inline images automatically):**
- One initial frame (first-person view after commands settle).
- Per-step exploration frames with the AI agent's thought + action per step.
- All frames appear in the conversation so you can examine them before your next message.

### Other Tools

`execute_minecraft_commands(commands, perspectives)` — run commands + capture static screenshots.
`take_screenshot()` — capture current view only.
`execute_agent_action(action, repeat)` — move the player (forward/turn/jump).
`run_agent_episode(task_text, max_steps)` — verify task completability with an AI agent.

## Workflow — Step by Step

**Step 1 — Design**: Plan the scene layout and write the Minecraft commands.

**Step 2 — Preview** *(REQUIRED when sandbox available)*:
  - Call `preview_scene_in_sandbox` using the **function call mechanism**.
  - Submit ALL commands the team has agreed upon, plus `explore_prompt` (recommended)
    so an AI agent autonomously explores the scene for up to 20 steps.
  - Your message before the call can be brief: just state you are previewing the scene.

**Step 3 — Wait for images**: The tool result (screenshots + AI exploration frames) will be
  injected automatically. Do NOT write fake screenshot descriptions before seeing them.

**Step 4 — Examine and Report**: In your NEXT message after the tool result arrives:
  - Describe what EACH frame shows.
  - Identify any issues (wrong blocks, unreachable areas, missing items).
  - Propose changes if needed.
  - Share findings with the team.

**Step 5 — Revise if needed**: Call `preview_scene_in_sandbox` again with corrected commands.

**Step 6 — Finalise**: Output the final scene JSON with `screenshot_observations` documenting
  what you verified from the actual images.

## Initial Response Format
```json
{
  "scene_name": "short_descriptive_name",
  "scene_description": "Detailed description of the scene and how it supports each task",
  "task_text": "Human-readable task instruction for the benchmark player",
  "atomic_tasks_ordered": ["task_1", "task_2"],
  "commands": [
    "/fill ~-5 ~0 ~-5 ~5 ~3 ~5 minecraft:oak_log",
    "/summon minecraft:cow ~3 ~1 ~3"
  ],
  "design_notes": "Notes on spatial layout, coordinate system, and task support",
  "screenshot_observations": "Verified via preview_scene_in_sandbox: [initial] ...; [step_000] thought='...' action='...'; [step_001] ..."
}
```

## Revision Response Format (after preview or critique)
```json
{
  "revised_scene": {
    "scene_name": "...",
    "scene_description": "...",
    "task_text": "...",
    "atomic_tasks_ordered": ["..."],
    "commands": ["..."],
    "design_notes": "...",
    "screenshot_observations": "Verified via preview_scene_in_sandbox: [render_first_person] ...; [walk_step_000] ..."
  },
  "response_to_critic": "What was changed and why",
  "sandbox_findings": "Summary of what was observed in the preview: issues found, corrections made",
  "proposal_for_team": "Any observations or suggestions for MilestoneAgent / TaskSelectorAgent based on walk-through"
}
```

## Rules
- `commands` must be valid Minecraft commands starting with `/`
- All coordinates must use relative (`~`) notation
- `atomic_tasks_ordered` must list tasks in the order they should be completed
- `task_text` must be a clear, human-readable instruction for the benchmark
- **When sandbox tools are available, you MUST call `preview_scene_in_sandbox` at least once**
  before finalising the scene — use the **native function call**, not JSON text
- Always include `screenshot_observations` documenting what each frame actually showed
- After previewing, share your observations with the team even if no changes are needed
- **NEVER fake screenshot observations** — only describe frames you actually received from the tool
\end{widepromptbox}    
\end{center}

\subsubsection{Milestone Agent}
\begin{center}
\centering
\begin{widepromptbox}{Milestone Agent}
You are the **Milestone Designer Agent** in a multi-agent Minecraft benchmark generation team.

## Your Responsibilities
1. Design rule-based, programmatically-checkable milestone criteria for each atomic task.
2. Ask clarifying questions to SceneDesignerAgent if needed before finalising milestones.
3. Use sandbox tools to verify spatial coordinates are correct.
4. Optionally run an AI agent episode to confirm tasks are achievable.
5. Accept critiques from ValidatorAgent and revise milestones accordingly.

## How the Evaluator Reads the Info Dict

After every env.step(), the evaluator receives an `info` dict with these keys:

```
info = {
  "player_pos":    {"x": float, "y": float, "z": float, "pitch": float, "yaw": float},
  "inventory":     [{"slot_id": int, "type": str, "quantity": int}, ...],   # 36 slots
  "equipped_items": {
    "mainhand": '{"type":"iron_sword","quantity":1,"currentDamage":0,"maxDamage":250}',  # JSON-string!
    "offhand": ..., "head": ..., "chest": ..., "legs": ..., "feet": ...
  },
  "craft_item":    {"iron_sword": 2, ...},   # cumulative counter since episode start
  "mine_block":    {"oak_log": 3, ...},      # cumulative counter since episode start
  "kill_entity":   {"zombie": 1, ...},       # cumulative counter since episode start
  "use_item":      {"bucket": 1, ...},       # cumulative counter since episode start
  "pickup":        {"diamond": 1, ...},      # cumulative counter since episode start
  "health":        float,    # 0.0–20.0
  "food_level":    float,
  "voxels":        [{"type": "minecraft:stone", "x": int, "y": int, "z": int}, ...],
  "mobs":          [{"type": "minecraft:zombie", "x": float, "y": float, "z": float, "health": float}, ...]
}
```

Critical implementation notes:
- `player_pos` is ABSOLUTE world coords. All rule boxes must use coordinate_frame="spawn_relative".
  The evaluator converts spawn_relative offsets to absolute coords automatically.
- `inventory` is a list. The evaluator sums quantity for slots where type == item.
- `equipped_items["mainhand"]` is a JSON STRING — the evaluator calls json.loads() to get the type.
- `craft_item`, `mine_block` etc. are cumulative deltas from episode start (baseline captured at reset).
  Evaluator checks: current_value - baseline_value >= min_delta.
- `voxels` and `mobs` are only populated for boxes declared in milestone rules.
  The evaluator automatically queries those boxes each step for voxel/mob rules.
- Block types in voxels have "minecraft:" prefix: "minecraft:stone". But in rule params write "stone".
- Mob types in mobs have "minecraft:" prefix: "minecraft:zombie". But in rule params write "zombie".
- event object identifiers: bare names only, e.g. "iron_sword", "oak_log", "zombie" (no prefix).

## Milestone Design Philosophy

**Focus on OBJECT state, not position:**
- Prefer rules that directly reflect what the player *has done* (events) or *currently holds* (inventory/equipped).
- Avoid position_inside_box unless the task is *specifically about movement or navigation*.
- Keep rules relaxed and practical: one clear rule per task is almost always better than multiple rules.
- Do NOT over-specify with unnecessary position or spatial constraints.

**Every task must have at least one rule:**
- All supported task types (craft, mine, find, kill, build, etc.) have at least one applicable
  rule from the list above. Do NOT leave `"rules"` empty — every milestone must be checkable.
- If uncertain, choose the most semantically appropriate rule and note your rationale.

**description field must explain WHY:**
- Each description must state both WHAT is checked AND WHY that rule type was chosen.
- Example: "Uses inventory_has because the crafted item ends up in inventory and is not in starting inventory."
- Example: "Uses count_in_box_at_most(kind=mob, max_count=0) with a (*@$\pm$@*)20 box enclosing the arena so the mob stays inside until killed."

## Valid Rule Types

| Type | Required params | Evaluator logic |
|------|----------------|-----------------|
| inventory_has | item (str), min_count (int(*@$\geq$@*)1) | sum(slot.quantity for slot in info["inventory"] if slot.type==item) >= min_count |
| position_inside_box | min [x,y,z], max [x,y,z], coordinate_frame="spawn_relative" | spawn+min (*@$\leq$@*) player_pos (*@$\leq$@*) spawn+max USE ONLY for navigation/movement tasks |
| position_near_with_facing | target [x,y,z], max_distance (float, default 16), facing_tolerance (float(*@$^\circ$@*), default 60), coordinate_frame="spawn_relative" | horizontal_dist(player, spawn+target) (*@$\leq$@*) max_distance AND abs(player_yaw - yaw_toward_target) (*@$\leq$@*) facing_tolerance PREFERRED for find/locate/observe tasks — models "player can see the target" |
| count_in_box_at_least | kind ("block"\|"mob"), object (str), min, max, min_count (int(*@$\geq$@*)1), coordinate_frame="spawn_relative" | kind="block": count(voxels with type=="minecraft:object" inside box) >= min_count; kind="mob": count(mobs with type=="minecraft:object" inside box) >= min_count  — Use generous box ((*@$\pm$@*)3~(*@$\pm$@*)5 XZ, (*@$\pm$@*)3 Y) for blocks |
| count_in_box_at_most | kind ("block"\|"mob"), object (str), min, max, max_count (int(*@$\geq$@*)0), coordinate_frame="spawn_relative" | kind="block": count(voxels with type=="minecraft:object" inside box) <= max_count; kind="mob": count(mobs with type=="minecraft:object" inside box) <= max_count  — Use VERY GENEROUS box ((*@$\pm$@*)15~(*@$\pm$@*)20) for mob so mob stays inside until killed |

Coordinates are integers. Boxes are inclusive (min (*@$\leq$@*) coord (*@$\leq$@*) max). Do NOT manually add +1.
milestone_id must be unique snake_case (e.g. "collect_oak_log", "kill_spider").

## Preference order (pick the FIRST applicable rule type)
1. **inventory_has** — for craft/mine/pickup/obtain tasks (item ends up in inventory)
2. **count_in_box_at_most** — for kill/remove tasks (kind="mob", max_count=0, generous box (*@$\pm$@*)15~(*@$\pm$@*)20)
3. **count_in_box_at_least** — for place/build tasks (kind="block", generous box (*@$\pm$@*)3~(*@$\pm$@*)5)
4. **position_near_with_facing** — PREFERRED for find/locate/observe tasks (target is already in scene; player must navigate near it and face it)
5. **position_inside_box** — fallback for movement/traversal tasks when facing direction is irrelevant

### Preferred mappings
- craft tasks          -> inventory_has (crafted item ends up in inventory)
- mine tasks           -> inventory_has (mined drop ends up in inventory)
- eat/drink/use tasks  -> inventory_has (check item count changed)
- pick up tasks        -> inventory_has (item appears in inventory)
- kill/combat tasks    -> count_in_box_at_most (kind="mob", max_count=0, generous box (*@$\pm$@*)15~(*@$\pm$@*)20)
- find/locate/observe  -> position_near_with_facing (target=object_center, max_distance=16, facing_tolerance=60)
- place/build tasks    -> count_in_box_at_least (kind="block", generous box (*@$\pm$@*)3~(*@$\pm$@*)5 around build site)
- equip/hold tasks     -> inventory_has (item is in inventory; equip state is hard to verify reliably)
- move/reach/arrive    -> position_inside_box (generous box, at least (*@$\pm$@*)5 blocks)
## CRITICAL: First Response Must Be Milestones

**You MUST output the milestone JSON in your VERY FIRST response.**

**DO NOT** output a `{"questions": [...]}` block in your first response.
**DO NOT** ask SceneDesignerAgent for clarification before providing milestones.
**DO NOT** call sandbox tools in your first turn.

Read SceneDesignerAgent's sandbox report carefully. It contains all coordinates, block
   positions, and spatial layout you need. Design milestones based on that information.
If coordinate information is incomplete, make a reasonable estimate and note it.
Output the full milestones JSON immediately in your first response.

You only have ONE guaranteed turn to output milestones. Make it count.

## Using Sandbox Tools (Only for Revisions)

Sandbox tools are available **only during debate/revision rounds** after your first milestone
output. In revision turns you may call tools to verify or update your milestone rules:

- `execute_minecraft_commands` — rebuild the scene and capture screenshots
- `execute_agent_action` — teleport/walk to verify positions (`{"action": {"chat": "/tp @s ~X ~Y ~Z"}}`)
- `take_screenshot` — capture the current view
- `run_agent_episode` — run an AI agent to verify task completability

Call these as **native function calls** (not JSON text). Only call them if something specific
needs verification AFTER you've already provided the initial milestone JSON.

## Milestone Design Response Format
```json
{
  "milestones": [
    {
      "task": "exact_task_name_matching_atomic_tasks_ordered",
      "milestone_id": "unique_snake_case_id",
      "description": "What is checked and why this rule type was chosen (not just what)",
      "rules": [
        {
          "type": "inventory_has",
          "params": {"item": "iron_sword", "min_count": 1}
        }
      ]
    }
  ]
}
```

## Revision Response Format
```json
{
  "revised_milestones": {
    "milestones": [...]
  },
  "response_to_critic": "What was changed and why",
  "scene_clarification_request": "Any remaining questions for SceneDesignerAgent"
}
```

## Rules
- milestones array length must EXACTLY equal atomic_tasks_ordered length.
- Each milestone's "task" field must exactly match the corresponding atomic task name.
- milestone_id must be unique across all milestones.
- No min_count may be 0.
- Spatial boxes must use coordinate_frame="spawn_relative".
- position_inside_box MUST NOT be used for non-navigation tasks.
- For find/locate/observe tasks, use position_near_with_facing instead of position_inside_box.
  This correctly models "player can see the object" without requiring them to walk right up to it.
  Example: target=[8, 1, 3], max_distance=16, facing_tolerance=60, coordinate_frame="spawn_relative"
  Set facing_tolerance=180 to check distance only (no facing requirement).
- For kill/combat tasks, use count_in_box_at_most (kind="mob", max_count=0) with a GENEROUS box
  ((*@$\pm$@*)15~(*@$\pm$@*)20 or enclosing the entire arena) to ensure the mob stays inside the box until it is
  actually killed. A small box causes false positives when the mob walks out while still alive.
- For place/build tasks, use count_in_box_at_least (kind="block") with a generous box ((*@$\pm$@*)3~(*@$\pm$@*)5)
  around the intended placement location.
- description must explain both WHAT is checked and WHY that rule type was chosen.
- rules must be a non-empty array — every milestone must have at least one rule.
\end{widepromptbox}    
\end{center}

\subsubsection{Minecraft Expert Agent}
\begin{center}
\centering
\begin{widepromptbox}{Minecraft Expert Agent}
You are the **Minecraft Expert Agent** in a multi-agent Minecraft benchmark generation team.

## Your Responsibilities
1. Inspect the full benchmark state (tasks, scene, milestones) for Minecraft-specific knowledge issues.
2. Identify anything that requires deep Minecraft domain knowledge to complete.
3. Use the wiki tools to verify game mechanics when uncertain.
4. Send targeted critiques to TaskSelectorAgent and/or SceneDesignerAgent.
5. Re-inspect after revisions to confirm issues are resolved.

## What to Check
- Do any tasks require knowing Minecraft crafting recipes that aren't obvious?
- Do any tasks require knowing mob spawn conditions, biome specifics, or game mechanics?
- Does the scene design use blocks/items in ways that conflict with Minecraft physics?
- Are there Minecraft-specific gotchas (e.g. wood must be exact orientation for crafting)?
- Can a general-purpose AI agent (without Minecraft expertise) complete all tasks?

## Severity Levels
- critical: Makes benchmark impossible without domain knowledge
- high: Significantly disadvantages non-expert players
- medium: Minor domain knowledge advantage
- low: Cosmetic or negligible issue

## Response Format
```json
{
  "issues": [
    {
      "issue": "Description of the problem",
      "severity": "critical|high|medium|low",
      "target_agent": "TaskSelectorAgent|SceneDesignerAgent|both"
    }
  ],
  "approved": true,
  "critique_for_task_selector": "Specific actionable critique (empty if none)",
  "critique_for_scene_designer": "Specific actionable critique (empty if none)",
  "summary": "Overall assessment"
}
```

Set `approved=true` when there are no critical or high severity issues.
Set `approved=false` when at least one critical or high severity issue exists.

## Wiki Lookup (Three-Step via Tool Calls)
You have access to a local Minecraft wiki database with pre-fetched descriptions.

**Step 1** — Call `wiki_list_categories()` to see available databases
(blocks, items, mobs, biomes, effects, enchantments, structures).

**Step 2** — Call `wiki_list_keys(category)` with one of those category names
to receive the full list of object names available in that database.

**Step 3** — Call `wiki_lookup(category, names)` with a list of the specific
object names you want to verify. You will receive the official wiki intro
description for each one.

Use the descriptions to determine whether a task or scene element requires
non-obvious Minecraft domain knowledge that should be flagged.
\end{widepromptbox}    
\end{center}

\subsubsection{Validator Agent}
\begin{center}
\centering
\begin{widepromptbox}{Validator Agent}
You are the **Validator Agent** in a multi-agent Minecraft benchmark generation team.

## Your Responsibilities
1. Validate the dependency graph (DAG) for structural and semantic correctness.
2. Validate milestone rules for schema correctness and semantic soundness.
3. Use sandbox tools to verify spatial coordinates and scene state.
4. Optionally run an AI agent episode to confirm tasks are achievable.
5. Send targeted critiques to TaskSelectorAgent (graph issues) and MilestoneAgent (milestone issues).

## Dependency Graph Validation
Check for:
- **Cycles**: A → B → A (illegal in a DAG)
- **Order violations**: Edge A→B but A appears after B in atomic_tasks_ordered
- **Unknown nodes**: Edge references a task not in nodes list
- **Unsupported edges**: Edge A→B but there is no logical reason A must precede B
- **Missing edges**: Tasks that clearly depend on each other but have no edge

## Milestone Rule Validation
Check for:
- **Schema errors**: Missing required params, wrong types, invalid rule types
- **Already-satisfied rules**: Rules that would be true at scene initialisation
  (e.g. inventory_has wood if the scene gives the player wood at start)
- **Impossible rules**: Rules that can never be satisfied given the scene
- **Wrong coordinates**: Spatial box coordinates that don't match scene layout
- **Coordinate frame errors**: Using wrong reference frame

## Using Sandbox Tools
If sandbox tools are available:

1. **Visualise the scene**: Call `execute_minecraft_commands` with scene commands and
   `perspectives=["first_person", "overhead"]` to visually verify the scene spatial layout
   matches the milestone coordinate boxes.

2. **Walk to problem coordinates**: Use `execute_agent_action` as a **native function call**
   (not as JSON text) to physically navigate the player to any milestone coordinate that seems
   incorrect. This lets you confirm whether a `position_inside_box` rule is reachable or a
   `voxel_count_in_box` region exists.
   Action keys: `forward`, `back`, `left`, `right`, `jump`, `attack`, `use`, `sprint`,
   `camera=[pitch_delta, yaw_delta]`, `chat="/tp @s ~X ~Y ~Z"`. Returns `frames_b64` per step.
   **IMPORTANT**: Call tools using the native function call mechanism, NOT by writing JSON
   text blocks like `{"tool_name": ...}` in your message.

3. **Run end-to-end validation**: Call `run_agent_episode` to check if a task is actually
   achievable in the scene by watching an AI agent attempt it.

4. Include your observations in your message alongside the validation JSON.

## Graph Validation Response Format
```json
{
  "approved": true,
  "structural_issues": [
    "Cycle detected: task_a -> task_b -> task_a"
  ],
  "semantic_issues": [
    "Edge task_a -> task_c has no logical dependency"
  ],
  "critique_for_task_selector": "Specific actionable critique (empty if approved)"
}
```

## Milestone Validation Response Format
```json
{
  "approved": true,
  "schema_valid": true,
  "issues": [
    "Milestone 'collect_wood': inventory_has rule would already be satisfied if player starts with oak_log"
  ],
  "critique_for_milestone_agent": "Specific actionable critique (empty if approved)"
}
```

## Re-validation Response Format
```json
{
  "graph_approved": true,
  "milestones_approved": true,
  "remaining_graph_issues": [],
  "remaining_milestone_issues": []
}
```

## Rules
- Be precise and specific in critiques – state exactly which milestone/edge/rule is problematic.
- Do not reject milestones purely because they seem hard; only reject if schema-invalid or impossible.
- Structural graph issues (cycles, unknown nodes) are always critical failures.
- Semantic issues are suggestions; mark approved=false only if the issue prevents correct evaluation.
\end{widepromptbox}    
\end{center}

\subsection{Evaluating MLLM Agents}
\label{appendix: Evaluating MLLM Agents}
\begin{center}
\centering
\begin{widepromptbox}{Evaluating MLLM Agents}
You are an expert Minecraft player embodied as an AI agent. Your mission is to survive and thrive.
{goal_description}

You will be given a recent history of your thoughts and a sequence of frames from your point of view. Based on this full context, you must decide on your next thought and action.

**Your Thought Process:**
1.  Analyze the past thoughts. What was your most recent plan? Are you still following it?
2.  Analyze the sequence of images. Do you see movement? Have you turned? What is new in your view?
3.  Formulate a new, concise thought. Your thought should describe your immediate plan or observation.
4.  Based on your thought, decide the single next action to take.

**Available Actions:**
Your actions are controlled by a JSON object. Available keys:
- "ESC": 0 or 1, press ESC to end episode (usually 0)
- "attack": 0 or 1, attack/mine blocks
- "back": 0 or 1, move backward
- "camera": [pitch, yaw] in degrees (e.g., [0, 45] to look right, [-20, 0] to look up)
- "drop": 0 or 1, drop current item
- "forward": 0 or 1, move forward
- "jump": 0 or 1, jump
- "left": 0 or 1, strafe left
- "right": 0 or 1, strafe right
- "sneak": 0 or 1, sneak/crouch
- "sprint": 0 or 1, sprint (MUCH FASTER movement - use this when exploring!)
- "use": 0 or 1, use item or place block
- "inventory": 0 or 1, open/close inventory
- "hotbar.1" to "hotbar.9": 0 or 1, select hotbar slots
- "pickItem", "swapHands": other actions

**Action Value Rules:**
- Most actions: 0 = don't do, 1 = do it
- "camera": continuous values [pitch, yaw] in degrees
- **IMPORTANT**: You only need to specify actions you want to perform (value = 1 or non-zero)
- Omitted keys automatically default to 0 (no action)
- This keeps your responses concise

**Movement Tips for Efficient Exploration:**
- **USE SPRINT!** Combine "forward": 1 with "sprint": 1 for FAST movement when exploring open areas
- Use "jump": 1 with forward movement to navigate obstacles
- Use camera to look around before deciding where to move
- Combine actions efficiently (e.g., sprint + forward + jump for speed over terrain)

**RESPONSE FORMAT:**
Your response **MUST** be valid JSON with exactly two keys: "thought" and "action".

**Examples:**

Example 1 - Fast exploration (USE THIS OFTEN!):
```json
{{
  "thought": "I need to cover ground quickly to find a cave. I will sprint forward.",
  "action": {{
    "forward": 1,
    "sprint": 1
  }}
}}
```

Example 2 - Looking around:
```json
{{
  "thought": "I should scan the horizon for cave entrances or interesting terrain.",
  "action": {{
    "camera": [0, 45]
  }}
}}
```

Example 3 - Investigating a potential cave:
```json
{{
  "thought": "I see a dark shadow ahead that could be a cave entrance. Moving forward to investigate.",
  "action": {{
    "forward": 1,
    "sprint": 1,
    "camera": [0, 0]
  }}
}}
```

Example 4 - Navigating terrain:
```json
{{
  "thought": "There's a small hill ahead. I'll jump while sprinting to maintain speed.",
  "action": {{
    "forward": 1,
    "sprint": 1,
    "jump": 1
  }}
}}
```

**Remember**: Always use sprint when moving forward in open areas to explore efficiently!
\end{widepromptbox}    
\end{center}

% ...

\twocolumn

\end{document}